\documentclass[10pt,twocolumn,letterpaper]{article}

\usepackage{wacv}
\usepackage{times}
\usepackage{epsfig}
\usepackage{graphicx}
\usepackage{amsmath}
\usepackage{amssymb}
\usepackage{multirow}
\usepackage{graphicx}
\usepackage{adjustbox}
\usepackage{color}
\usepackage{subfigure}
\usepackage{wrapfig}
\usepackage{lipsum}
\usepackage{caption}
\usepackage{float}
\usepackage{booktabs}
\usepackage[accsupp]{axessibility}  

\def\wacvPaperID{1281} 

\wacvfinalcopy 

\ifwacvfinal
\fi

\ifwacvfinal
\usepackage[breaklinks=true,bookmarks=false]{hyperref}
\else
\usepackage[pagebackref=true,breaklinks=true,colorlinks,bookmarks=false]{hyperref}
\fi

\begin{document}

\title{Fast and Explicit Neural View Synthesis}


\author{ 
Pengsheng Guo \quad Miguel Angel Bautista \quad Alex Colburn \quad Liang Yang \\
Daniel Ulbricht \quad Joshua M. Susskind \quad Qi Shan\\
Apple \\
{\tt\small \{pengsheng\_guo,mbautistamartin,alexcolburn,ericyang,dulbricht,jsusskind,qshan\}\href{mailto:pengsheng_guo@apple.com, qshan@apple.com}{@apple.com}} \\
}

\maketitle

\ifwacvfinal
\thispagestyle{empty}
\fi

\begin{abstract}
   
We study the problem of novel view synthesis from sparse source observations of a scene comprised of 3D objects. We propose a simple yet effective approach that is neither continuous nor implicit, challenging recent trends on view synthesis. 
Our approach explicitly encodes observations into a volumetric representation that enables amortized rendering.
We demonstrate that although continuous radiance field representations have gained a lot of attention due to their expressive power, our simple approach obtains comparable or even better novel view reconstruction quality comparing with state-of-the-art baselines \cite{pixelNeRF} while increasing rendering speed by over 400x. Our model is trained in a category-agnostic manner and does not require scene-specific optimization. Therefore, it is able to generalize novel view synthesis to object categories not seen during training. In addition, we show that with our simple formulation, we can use view synthesis as a self-supervision signal for efficient learning of 3D geometry without explicit 3D supervision.
\end{abstract}


\section{Introduction}

In order to understand the 3D world, an intelligent agent must be able to perform quick inferences about a scene's appearance and shape from unseen viewpoints given few observations. Being able to synthesize images at target camera viewpoints efficiently given sparse source views serves a fundamental purpose in building intelligent visual behaviour \cite{katerina1,katerina2,katerina3}. The problem of learning to synthesize novel views has been widely studied in literature, with approaches ranging from traditional small-baseline view synthesis relying on multi-plane imaging \cite{mpi1, mpi2, mpi3, llff}, flow estimation \cite{appearanceflow, multiview2novelview}, to explicitly modeling 3D geometry via point-clouds \cite{npbg}, meshes \cite{softraster}, and voxels \cite{enr}.

\begin{figure}[t]
\begin{center}
\vskip -0.5in
\centerline{\includegraphics[width=1.0\linewidth]{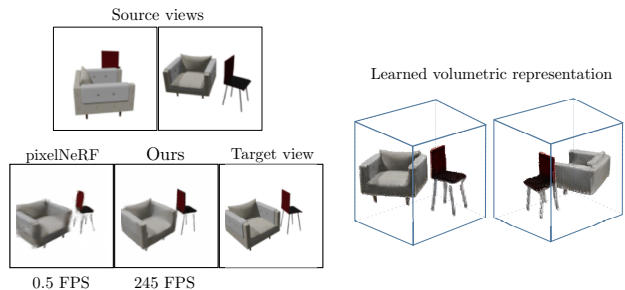}}
\vskip -0.55in
\caption{Our model performs scene-agnostic and category-agnostic novel view synthesis in real time. A complete 3D geometry of the scene is estimated through a single forward pass. Our model consistently produces higher quality results comparing to the state-of-the-art view synthesis approach pixelNeRF \cite{pixelNeRF}, while being over $400$x faster in rendering time.}
\label{fig:intro}
\vskip -0.3in
\end{center}
\end{figure}

A recent wave of approaches for view synthesis have adopted continuous radiance field representations \cite{pixelNeRF, srn,nerf,grf,nerfies}, where scenes are represented as a continuous function that shares its domain with the signal being fitted (\eg a function that takes points in $\mathbb{R}^3$ as input, to model a 3D signal), as opposed to discrete representations where the 3D signals are encoded in a discrete geometric structure like a volume \cite{enr} or a mesh \cite{softraster}.  Although continuous radiance field representations enjoy the benefits of being resolution-free or modeling view-dependent effects, they are not efficient for real-world use cases that require real-time performance. Typically, radiance field representations have the following disadvantages. First, being computationally costly to obtain when implicitly modeled \cite{nerf, nerfies, derf}, \eg the model parameters are optimized via gradient descent for each object or scene, usually taking tens of hours on commodity hardware. Second, requiring to densely capture observations of the scene being modeled \cite{nerf, nerfies, derf} for optimization. Third, not being able to amortize the rendering cost across views, since radiance fields are evaluated independently for every pixel being rendered \cite{pixelNeRF,grf}. This dramatically impacts the practicality of radiance field, since rendering an image can take seconds on modern GPUs.


What makes an approach for novel view synthesis useful? While photo-realistic results have been obtained with continuous/implicit representations, these approaches are severely impacted by capture, optimization and rendering time, hindering their practicality for deployment in systems that require real-time performance, \eg the ability to infer views of unseen objects in real-time from sparse observations. Our approach enjoys the following benefits: \textit{(i)} the scene representation is fast to obtain, as it does not require gradient-based optimization for new scenes and can be obtained from sparse observations, and \textit{(ii)} it is efficient to render, since it models the complete 3D geometry and appearance in a single forward pass, allowing for amortized rendering. Our experiments show that despite the simplicity of our method, our performance notably matches or beats recent state-of-the-art baselines based on few-shot continuous scene representations across different metrics and settings, producing accurate novel view reconstruction, while rendering objects over $400$x faster than the state-of-the-art, pixelNeRF \cite{pixelNeRF}. In addition, we find that the 3D geometry learned by our model in an unsupervised manner (i.e. without the need to train with 3D geometry supervision) is extremely compelling and very efficient to obtain, requiring only a single forward pass of the model.

\section{Related Work}

Learning to synthesize novel views of an object or a scene given one or more sparse observation has been widely studied in the literature \cite{mpi1,mpi2,mpi3,llff, extremeviewsynthesis,appearanceflow,monocularviewsyn,multiview2novelview,enr,tbn,npbg,srn}. A unifying problem definition for this set of approaches is to predict a target view given a source view/s, conditioned on a relative camera transformation. One set of approaches focuses on small and/or wide baseline view synthesis where the goal is to synthesize a parallax effect by using multi-plane imaging \cite{mpi1, mpi2, mpi3}, local light-field fusion \cite{llff} or cost volume estimation \cite{extremeviewsynthesis}. Another set of approaches focuses on learning a free-form 2D flow field that takes pixels from a single \cite{appearanceflow} or multiple source views \cite{multiview2novelview} and reconstructs a target view given the relative camera transformation between source/s and target.

In addition, there is an extensive literature on tackling view synthesis with voxel grid 3D representations \cite{enr, tbn, deepvoxels, von, hologan, platonicgan, neuralvolumes, tungcommonsense}. Although our approach uses a voxel grid 3D representation, it differs from existing work in the following. As opposed to \cite{enr, tbn, tungcommonsense} where convolutional 2D decoders are used to generate an image, our approach uses volumetric rendering to directly render an image from the explicit voxel grid representation. In contrast to \cite{deepvoxels, neuralvolumes}, our approach can generalize to multiple objects without per scene training/optimization. Moreover, our approach is trained in a category-agnostic way as opposed to \cite{von, hologan, platonicgan}, and it is trained on a large set of object categories (as opposed to 4 object categories in \cite{tungcommonsense}) which can be generalized to unseen object categories (cf. Sect. \ref{sec:experiments}).

In order to deal with the limitations of voxel grids, implicit representations that model continuous radiance fields for view interpolation \cite{nerf,nerfies,derf,sharf} have been proposed. These approaches learn a radiance field for every scene or object by fitting the parameters of a model (using gradient descent) to a dense set of views of a scene and then interpolating between those views. Note that this setting is different from the novel view synthesis setting where the problem is to predict a target view given a sparse source view and a relative camera transformations. However, recent approaches have applied continuous radiance fields to the novel view synthesis problem \cite{pixelNeRF, grf}, showing that it is possible to model multiple objects or scenes within a single model and extrapolating to object categories unseen during training. We can group recent approaches to novel view synthesis with implicit and continuous radiance field representations into two mutually exclusive categories. In the first category we find approaches that provide an efficient approach to explicitly encode source views into a continuous representation but are inefficient during rendering due not being able to amortize the rendering process across views \cite{pixelNeRF,grf,ibrnet} (see Sect. \ref{sec:rendering} for details). In the second category, we find recent approaches that enable efficient rendering through amortized rendering \cite{plenoctrees,fastnerf,nex} but where their continuous representation is implicit, and must be fitted via gradient descent for every new object or scene (typically taking days on commodity hardware). 

In this paper we present a simple yet powerful approach for novel view synthesis which explicitly encodes sources views into a volumetric representation that enables amortized rendering. Thus combining the best of both types of recent approaches for novel view synthesis.

\begin{figure*}[!t]
\begin{center}
\centerline{\includegraphics[width=0.85\linewidth]{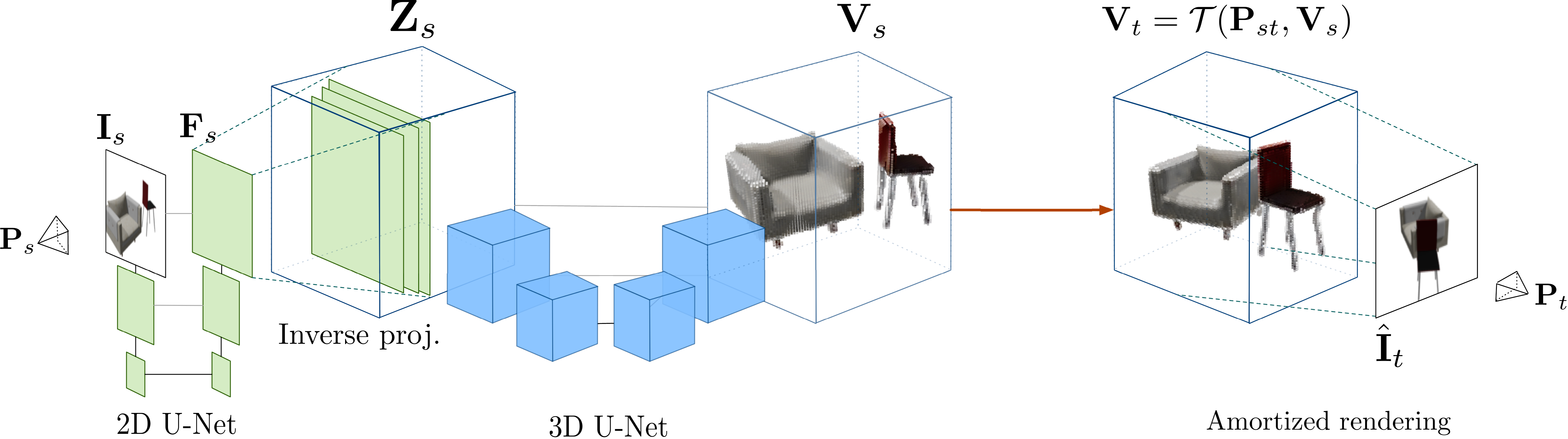}}
\caption{Our model is comprised of three main components: a) a 2D U-Net image encoder, b) a 3D U-Net scene encoder, and c) an amortized rendering process. The 2D U-Net encoder generates a 2D feature map $\mathbf{F}_s$ from the input image $\mathbf{I}_s$. The feature map is then projected into a latent volume $\mathbf{Z}_s$ via an inverse project step. A 3D U-Net network maps $\mathbf{Z}_s$ into an RGB$\alpha$ volume $\mathbf{V}_s$. This RGB$\alpha$ volume is applied a relative pose transformation $\mathbf{P}_{st}$ to match the target view pose $\mathbf{P}_t$, and the resulting image $\hat{\mathbf{I}}_t$ is created by rendering the RGB$\alpha$ using a simple volume rendering process that is amortized across views.}
\label{fig:arch}
\end{center}
\vskip -0.3in
\end{figure*}

\section{Methodology}
\label{sec:Methodology}

The novel view synthesis problem is defined as follows. Given a set $\mathcal{S} = \{ (\mathbf{I}_i, \mathbf{P}_i)\}_{i=0}^{n}$ of one or more source views, where a view is defined as an image $\mathbf{I}_i \in \mathbb{R}^{3 \times h \times w}$ together with the camera pose $\mathbf{P}_i \in SO(3)$, we want to learn a model $f_\theta$ that can reconstruct a ground-truth target image $\mathbf{I}_t$ conditioned on its pose $ \mathbf{P}_t$, where the predicted target image is obtained as $\hat{\mathbf{I}}_t = f_\theta(\mathcal{S}, \mathbf{P}_t)$.

We design $f_\theta$ as a simple fully convolutional model that allows amortized rendering. Our model processes a source view through a 2D U-Net encoder \cite{unet} to produce a feature map that is projected onto a latent volumetric representation via an inverse projection step. This volumetric representation is further processed with a 3D U-Net model to learn an RGB$\alpha$ volume\footnote{Note that we do not supervise training with an RGB$\alpha$ volume, the model is forced to learn the RGB$\alpha$ volume through the rendering process.} to which the relative pose transformation between source and target views is applied, and finally rendered into the predicted target view. We illustrate our pipeline in Fig. \ref{fig:arch}.

\subsection{Encoding}
\label{sec:encoding}
The initial step of our model is to encode the source $\mathbf{I}_s \in \mathbb{R}^{3 \times h \times w}$ with a fully convolutional U-Net encoder that produces a feature map $\mathbf{F}_s \in \mathbb{R}^{c \times h \times w}$ that preserves the spatial resolution of the source image. Once a feature map $\mathbf{F}_s$ is obtained, we cast the features along rays into a latent volumetric tensor using the perspective camera matrix. In practice we perform an inverse projection step to back-project $\mathbf{F}_s$ into a latent volumetric tensor $\mathbf{Z}_s \in \mathbb{R}^{c \times d_s \times h_s \times w_s}$, where $d_s, h_s, w_s$ are depth, height and width for the volumetric representation \footnote{We used intrinsic camera parameters for the inverse projection step, which we assume to be constant.}. 
Instead of reshaping 2D feature maps into a 3D volumetric representation \cite{enr, tbn}, we found that using an inverse projection step is beneficial to preserve the 3D geometry and texture information (cf. Sect. \ref{sec:experiments} for empirical evidence). 

\subsection{Learning a Renderable Volume}
After the inverse projection step we simply process $\mathbf{Z}_s$ with a 3D U-Net \cite{atlas} model and predict a final volume $\mathbf{V}_s \in \mathbb{R}^{4 \times d_s \times h \times w}$. At this stage $\mathbf{V}_s$ encodes an RGB$\alpha$ volume  of the object or scene that can be efficiently rendered. Similar to \cite{pixelNeRF,enr} we apply the relative transformation $\mathbf{P}_{st} = \mathbf{P}_t \mathbf{P}_{s}^{-1}$ between source and target camera poses to the volumetric representation $\mathbf{V}_s$ to obtain a transformed volumetric representation $\mathbf{V}_t$ that is aligned with the target view. We define this transformation operation as a function $\mathcal{T}(\mathbf{P}, \mathbf{V})$ that takes as input a rigid transformation $\mathbf{P} \in SO(3)$ and a volume $\mathbf{V}$ and applies the rigid transformation to the volume. Note that we define $\mathbf{P} \in SO(3)$, however, our formulation naturally extends to other transformations groups (\eg non-rigid or free-form deformations).

\subsection{Amortized Rendering}
\label{sec:rendering}

We now turn to the task of rendering an RGB$\alpha$ volume $\mathbf{V}$ into an image. Recent work on modeling scenes with continuous neural radiance fields \cite{nerf} has shown great results by using the rendering equation \cite{volumerendering} in order to model pixels. At rendering time, \cite{nerf} propose to obtain a pixel value by tracing the camera ray $\mathbf{r}$ from the near plane $t_n$ to the far plane $t_f$, and the expected color of a 2D pixel can be calculated as follows (see \cite{nerf} for details):

\begin{equation}
\label{eq:nerf_int}
\begin{aligned}
C(\mathbf{r}) = \int_{t_n}^{t_f} T(t) \sigma(\mathbf{r}(t)) \mathbf{c}(\mathbf{r}(t),\mathbf{d}) dt\\
\end{aligned}
\end{equation}

where $T(t)=\exp(-\int_{t_n}^{t_f} \sigma(\mathbf{r}(s)) ds)$ denotes the accumulated transmittance between the near plane and the current point $\mathbf{r}(t)$ along the ray. In practice, numerical quadrature and stratified sampling strategies are adopted to discretize the continuous integral and make the computation viable.

However, a critical problem of the sampling process in NeRF \cite{nerf} is that it prevents the rendering process to be \textit{amortized} across views. This is because each ray integral in Eq. \ref{eq:nerf_int} is independent and points sampled to approximate one ray integral are not reusable for other ray integrals in the scene. Our approach side-steps the need to perform sampling by modelling the scenes complete geometry and appearance as an RGB$\alpha$ volume $\mathbf{V} \in \mathbb{R}^{4\times d_s \times h \times w}$. This allows us to amortize rendering across views (since all rendered images of a scene share the same RGB$\alpha$ volume) obtaining dramatic rendering speed improvements without sacrificing reconstruction accuracy with respect to recent baselines \cite{pixelNeRF}.

Before rendering our RGB$\alpha$ volume $\mathbf{V}$, we apply a perspective deformation (using intrinsic camera parameters) on the viewing frustum using inverse warping and trilinear sampling \cite{stn} (see Appendix for details). For a given pixel location $(i, j)$, the expected color $\hat{C}$ is calculated as:
\begin{equation}
\hat{C}_{i,j} = \sum_{k=1}^{d_s} T_{i,j}^k \alpha_{i,j}^k \mathbf{c}_{i,j}^k \mbox{, where } T_{i,j}^k = \prod_{m=1}^{k-1} (1-\alpha_{i,j}^m)
\end{equation}

where $\mathbf{c}_{i,j}^k$ is the color value encoded in the first $3$ channels of $\mathbf{V}_t$ and $\alpha_{i,j}^k$ is the value at the last channel.

\subsection{Multiple View Aggregation}

Our model can take an arbitrary number of source views in $\mathcal{S}$ as input. In order to do so, we first obtain latent volumes $\mathbf{Z}_i$ for each source view $i$ using the same encoding process as in Sect. \ref{sec:encoding}. Next, we take an arbitrary source view $i^*$ in the set of source views as the origin of the coordinate system. Latent volumes $\mathbf{Z}_i$ are then aligned to this origin using the relative transformation $\mathbf{P}_{i, i^*}$ between corresponding pose $\mathbf{P}_i$ and origin pose $\mathbf{P}_{i^*}$. After that, we pool the aligned volumes by taking the mean across views:

\begin{equation}
\bar{\mathbf{Z}} = \frac{1}{n} \sum_{i\in n} \mathcal{T}(\mathbf{P}_{i, i^*}, \mathbf{Z}_i)
\end{equation}

Finally, the pooled volumetric latent $\bar{\mathbf{Z}}$ is fed to our 3D U-Net to generate an RGB$\alpha$ volume $\mathbf{V}$ which can be efficiently rendered as outlined in Sect. \ref{sec:rendering}.

\subsection{Training}

Similar to \cite{enr, pixelNeRF}, we sample tuples of source and target views together with their relative transformation $(\mathbf{I}_s, \mathbf{I}_t, \mathbf{P}_{st})$ during training. We use the model $f_\theta$ to predict the target from source $\hat{\mathbf{I}}_t = f_\theta(\mathbf{I}_s, \mathbf{P}_{st})$ and minimize a rendering loss. The rendering loss is a weighted sum of $\ell_2$ loss and SSIM \cite{ssim} loss, defined as

\begin{equation}
\begin{aligned}
\mathcal{L}_{\mathrm{render}} 
& = \sum_{t} \|f_\theta(\mathbf{I}_s, \mathbf{P}_{st}) - \mathbf{I}_t\|^2_2  \\
& + \lambda \mathcal{L}_{\mathrm{ssim}} (f_\theta(\mathbf{I}_s, \mathbf{P}_{st}), \mathbf{I}_t)
\end{aligned}
\end{equation}

One advantage of our formulation is that it supports the use of structural losses like SSIM \cite{ssim} during training. The SSIM loss has been previously proved useful for view synthesis \cite{enr}, and are not directly applicable to NeRF-like methods \cite{pixelNeRF}, as they randomly sample sparse rays from each image during training due to the computational constraints.

\section{Experiments}
\label{sec:experiments}

Our model is evaluated on a series of well established ShapeNet \footnote{licensed for non-commercial research purposes} \cite{shapenet} benchmarks where it achieves similar or better visual quality compared to the state-of-the-art method pixelNeRF \cite{pixelNeRF} and other recent baselines \cite{enr,srn}, while rendering objects in real time.  We also evaluate the 3D reconstruction capabilities of our model, where it outperforms baseline unsupervised 3D reconstruction models.  The following sections detail evaluations on category-specific view synthesis for scenes with single and multiple objects, as well as category-agnostic, multi-category, and unseen-category objects. 3D reconstruction is evaluated in Section \ref{sec:3D_recon}, and the design and effectiveness of different components of our model are discussed in Section \ref{sec:design_ablation}.

\subsection{Novel View Synthesis}
\label{sec:view_synthesis}

In the novel view synthesis experiments we compare our approach with several state-of-the-art techniques: ENR\cite{enr}, pixelNeRF \cite{pixelNeRF}, DVR \cite{dvr}, and SRN \cite{srn}. Our SSIM, PSNR, and LPIPS \cite{lpips} scores demonstrates that we produce comparable or better rendering quality than pixelNeRF \cite{pixelNeRF}  with an explicit volumetric scene representation, while increasing the rendering speed $100\times$ per view, allowing us to render scenes in real-time as show in Table \ref{tb:render_inference_perf_comparison}.

\subsubsection{Category-Specific View Synthesis of Single Objects}
We evaluate our model on the ShapeNet chairs and cars categories in single-view and two-view settings, following the same experimental protocol as baseline methods \cite{srn, enr, pixelNeRF}. These category-specific datasets contain $6,591$ different chairs and $3,514$ different cars. Each object has 50 views sampled uniformly on the full sphere, rendering images resolution $128 \times 128$ pixels. 

Following pixelNeRF, we train a single model for both the single-view and two-view settings. During training, we randomly choose either one or two source views to predict the target view. For evaluation, we use either one or two source views of an unseen object and predict 250 target views. Additionally, we also report the rendering time comparison between pixelNeRF and our method. 

\begin{figure*}[!t]
\begin{center}
\centerline{\includegraphics[width=1.0\textwidth]{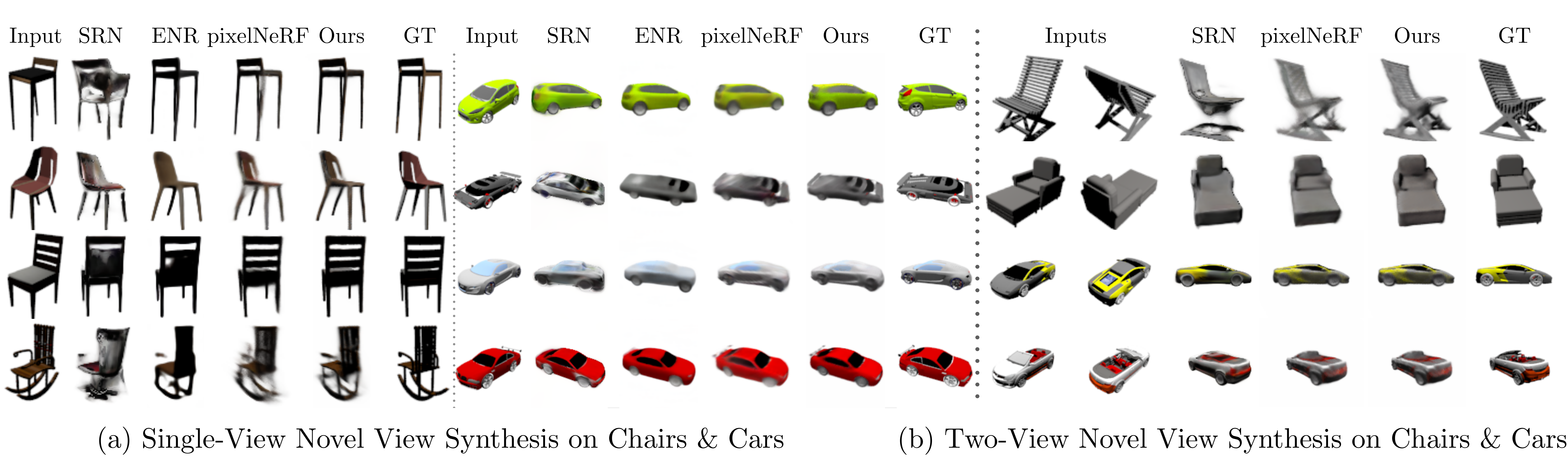}}
\vskip -0.1in
\caption{\textbf{Qualitative results on category-specific single chair \& single car.} The model can either take (a) single view or (b) two views as input to synthesis novel views. We find similar rendering quality comparing to pixelNeRF \cite{pixelNeRF} and better geometry prediction comparing to ENR \cite{enr} and SRN \cite{srn}. }
\label{fig:srn}
\end{center}
\end{figure*}

Despite its simplicity, our model obtains very competitive results compared to pixelNeRF, as shown in Table \ref{tb:SRN_datasets}. In general, we don't observe obvious mistakes made by our model when visually inspecting results. Fig. \ref{fig:srn} shows a random subset of source and predicted targets. 

In Table \ref{tb:render_inference_perf_comparison} we show the average inference and rendering time of both pixelNeRF and our approach.
The inference time is defined as the interval of time required to generate scene information (2D feature maps for pixelNeRF and 3D feature maps for our model) from the source views. The rendering time is the time required to render a target view given scene information. We compute per-view rendering time and per-object rendering time, where per-object rendering time is accumulated by rendering a total of $250$ views. To conduct a fair comparison, we equate the effective image batch size between pixelNeRF and our model. All the run times are reported on an NVIDIA Tesla V100 GPU. As shown in Table \ref{tb:render_inference_perf_comparison}, our per-view rendering time is $0.0178s$, $100$x faster than pixelNeRF, taking $1.9047s$ to render an image. In other words, our model achieves a rendering speed of $56$ FPS, which enables a real-time rendering experience. By amortizing the rendering step across multiple views, our model renders 250 views in $1.022s$ ($245$ FPS), while pixelNeRF renders the same 250 views in $474.8606s$ ($0.5$ FPS). This translates to over a $400$x speedup. In addition, we test the generalization capabilities of our model on real world data. We use the model trained with ShapeNet cars categories and perform novel view synthesis on real car images from \cite{realcar}. We found our model can generate plausible novel views with less artifacts and blurry effects compared to pixelNeRF \cite{pixelNeRF}. The complete experiments protocol and qualitative visualizations can be found in the appendix.

\begin{table}[ht]
\caption{Results on category-specific novel view synthesis for ShapeNet chairs and cars. Our method achieves competitive results compared to state-of-the-art approaches.}
\vskip -0.1in
\label{tb:SRN_datasets}
\begin{center}
\begin{adjustbox}{width=1\linewidth}
\begin{small}
{
    \begin{tabular}{cccccccccc}
    \toprule
    \multirow{2}{*}{Data} &\multirow{2}{*}{Methods} & \multicolumn{2}{c}{1-view} & \multicolumn{2}{c}{2-view} \\
    \cmidrule(lr){3-4}
    \cmidrule(lr){5-6}
     &  & PSNR$\uparrow$ & SSIM$\uparrow$ & PSNR$\uparrow$ & SSIM$\uparrow$\\
    \midrule
    \multirow{4}{*}{Chairs}
    & ENR & 22.83 & - & - & -\\
    & SRN & 22.89 & 0.89 & 24.48 & 0.92\\
    & pixelNeRF & \textbf{23.72} & 0.91 & \textbf{26.20}& \textbf{0.94}\\
    & Ours & 23.21 & \textbf{0.92} & 25.25 & \textbf{0.94}\\
    \midrule
    \multirow{4}{*}{Cars}
    & ENR & 22.26 & - & - & -\\
    & SRN & 22.25 & 0.89 & 24.84 & 0.92\\
    & pixelNeRF & \textbf{23.17} & 0.90 & \textbf{25.66} & \textbf{0.94}\\
    & Ours & 22.83 & \textbf{0.91} & 24.64 & 0.93\\
    \bottomrule
    \end{tabular}
}
\end{small}
\end{adjustbox}
\end{center}
\vskip -0.2in
\end{table}

\begin{table}[ht]
    \caption{Results on category-specific novel view synthesis for multiple chairs. Compared to pixelNeRF, our method predicts much more coherent synthesis results, and it beats pixelNeRF by a significant margin on all three metrics.}
    \vskip -0.1in
    \label{tb:multi_chair}
    \begin{center}
    \begin{small}
    {
            \begin{tabular}{cccccccccc}
        \toprule
        \multirow{2}{*}{Methods} & \multicolumn{3}{c}{2-view} \\
        \cmidrule(lr){2-4}
         & PSNR$\uparrow$ & SSIM$\uparrow$ & LPIPS $\downarrow$\\
        \midrule
        SRN & 14.67 & 0.664 & 0.431\\
        pixelNeRF & 23.40 & 0.832 & 0.207\\
        Ours & \textbf{24.13} & \textbf{0.907} & \textbf{0.098}\\
        \bottomrule
        \end{tabular}
    }
    \end{small}
    \end{center}
    \vskip -0.2in
    
\end{table}

\begin{table}[ht]
    \caption{Inference and rendering time (in seconds) analysis between pixelNeRF and our method. We show our model can achieve over $100$x faster  per-frame and over $400$x faster per-object rendering speed. }
    \vskip -0.1in
    \label{tb:render_inference_perf_comparison}
    
    \begin{center}
    \begin{adjustbox}{width=1\linewidth}
    \begin{small}
    {
        \begin{tabular}{cccccc}
        \toprule
        \multirow{2}{*}{} & \multicolumn{2}{c}{pixelNeRF} & \multicolumn{2}{c}{Ours}\\
        \cmidrule(lr){2-5}
         & Inference & Rendering & Inference & Rendering\\
        \midrule
        Per-view & 0.0053 & 1.8994 & 0.0146 & 0.0032\\
        Per-object & 0.0053 & 474.8553  & 0.0146 & 1.0074\\
        \bottomrule
        \end{tabular}
    }
    \end{small}
    \end{adjustbox}
\end{center}
    \vskip -0.2in
\end{table}

\subsubsection{Category-Specific View Synthesis of Multiple Objects}
We further extend the category-specific evaluation to the multiple-chair dataset proposed by pixelNeRF. This dataset consists of images rendered with two randomly located and oriented chairs. The dataset is designed so that the model cannot simply rely on certain semantic cues such as the symmetric property of a chair to perform geometry completion. The learned model should be flexible and robust enough to represent scenes instead of a single object. All images are rendered with a resolution of $128 \times 128$.

\begin{figure*}[!t]
\begin{center}
\centerline{\includegraphics[width=1.0\textwidth]{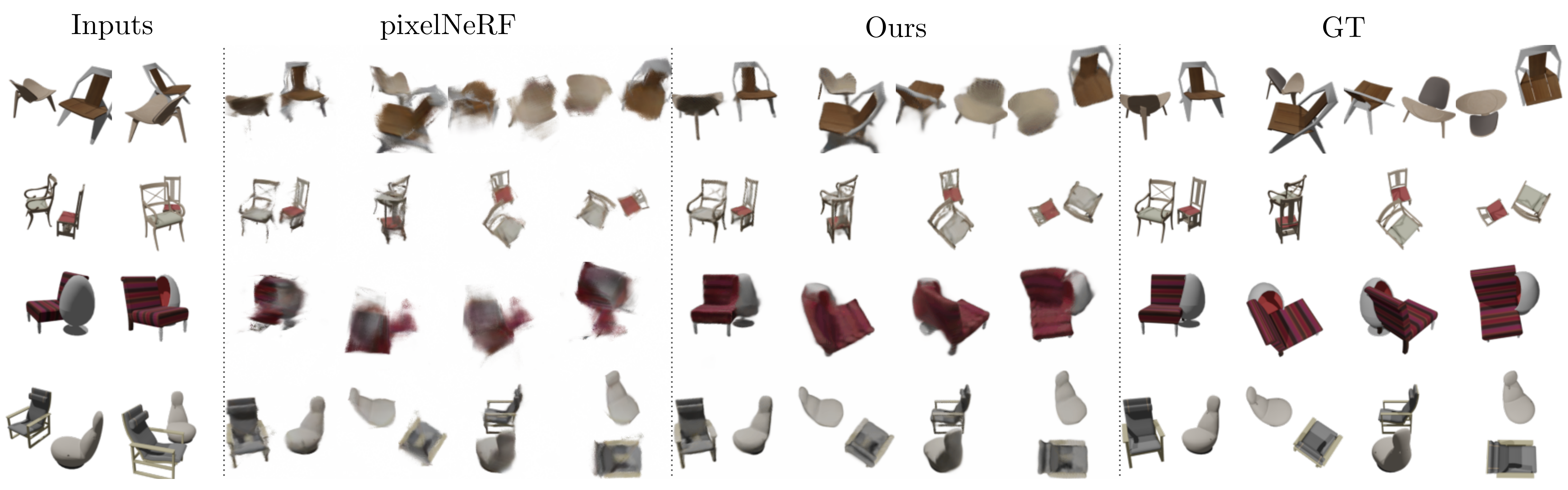}}
\caption{\textbf{Qualitative results on category-specific multiple chairs.} The models take two-view images as input. Compared to pixelNeRF, our model renders a cleaner appearance and more complete geometry for chairs with complex shapes.}
\label{fig:multi_chair}
\end{center}
\end{figure*}

We report reconstruction quality metrics in Table \ref{tb:multi_chair}. Despite the increased complexity of this setting, our simple model outperforms pixelNeRF across metrics, and exceeds the object-centric method SRN \cite{srn} by a large margin. Fig. \ref{fig:multi_chair} shows randomly sampled qualitative results. We observe that the views rendered by our model have cleaner geometry than pixelNeRF, which fails to predict a reasonable geometry at certain angles and suffers from ghosting artifacts.

\begin{figure*}[!t]
\begin{center}
\centerline{\includegraphics[width=0.95\linewidth]{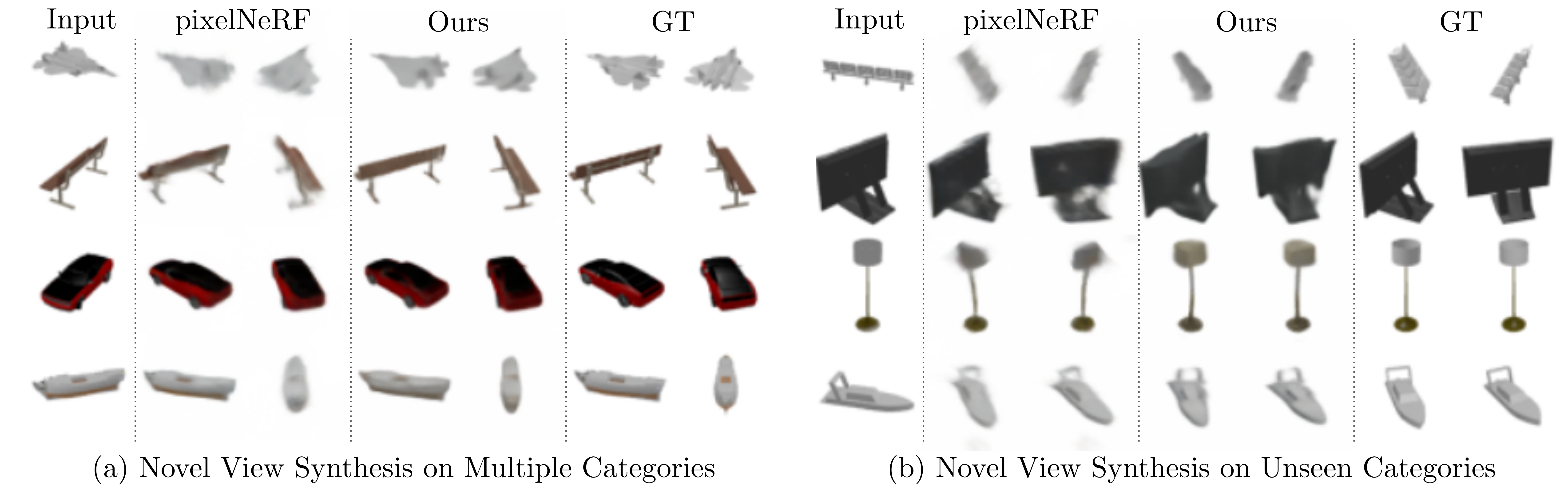}}
\caption{\textbf{Qualitative results on (a) category-agnostic and (b) unseen-category datasets.} We test the capacity of our model by training it across different categories in a single-view setting. We evaluate the performance on both seen an unseen categories. We consistently observe cleaner views predicted by our model compared to the baseline.}
\label{fig:multi_cat_unseen}
\end{center}
\vskip -0.3in
\end{figure*}

\subsubsection{Category-agnostic View Synthesis}

The category-agnostic setting is much more challenging than the category-specific one, because the model needs capacity to jointly learn objects across a range of completely different categories. To evaluate our model in the  category-agnostic setting, we follow the same training protocol as baseline method \cite{nmr} and evaluate on 13 different categories. Each object was rendered in 24 different views with a resolution of $64 \times 64$. We summarize our results in Table \ref{tb:NMR_datasets}. Our model beats all baseline methods in every metric. The qualitative visualization in Fig. \ref{fig:multi_cat_unseen} indicates our model can generate more clean geometry compared to pixelNeRF, which is corroborates the results obtained by our method in the multi-chair dataset. 

\begin{table*}[!t]
\caption{\textbf{Quantitative results on category-agnostic view synthesis.} Our model beats all baselines with a noticeable margin in terms of the mean metrics. The LPIPS score for our mode is significant better compared to state-of-the-art methods in all categories.}
\vskip -0.2in
\label{tb:NMR_datasets}
\begin{center}
\begin{adjustbox}{width=1\linewidth}
\begin{small}
{
\begin{tabular}{cccccccccccccccc}
\toprule
Metrics & Methods & plane & bench & cbnt. & car & chair & disp. & lamp & spkr. & rifle & sofa & table & phone & boat & mean\\
\midrule
\multirow{4}{*}{PSNR$\uparrow$}
& DVR & 25.29 & 22.64 & 24.47 & 23.95 & 19.91 & 20.86 & 23.27 & 20.78 & 23.44 & 23.35 & 21.53 & 24.18 & 25.09 & 22.70\\
& SRN & 26.62 & 22.20 & 23.42 & 24.40 & 21.85 & 19.07 & 22.17 & 21.04 & 24.95 & 23.65 & 22.45 & 20.87 & 25.86 & 23.28\\
& pixelNeRF & 29.76 & 26.35 & 27.72 & 27.58 & 23.84 & \textbf{24.22} & \textbf{28.58} & 24.44 & \textbf{30.60} & 26.94 & 25.59 & 27.13 & \textbf{29.18} & 26.80\\
& Ours       & \textbf{30.15} & \textbf{27.01} & \textbf{28.77} & \textbf{27.74} & \textbf{24.13} & 24.13 & 28.19 & \textbf{24.85} & 30.23 & \textbf{27.32} & \textbf{26.18} & \textbf{27.25} & 28.91 & \textbf{27.08} \\
\midrule
\multirow{4}{*}{SSIM$\uparrow$}
& DVR & 0.905 & 0.866 & 0.877 & 0.909 & 0.787 & 0.814 & 0.849 & 0.798 & 0.916 & 0.868 & 0.840 & 0.892 & 0.902 & 0.860\\
& SRN & 0.901 & 0.837 & 0.831 & 0.897 & 0.814 & 0.744 & 0.801 & 0.779 & 0.913 & 0.851 & 0.828 & 0.811 & 0.898 & 0.849\\
& pixelNeRF & 0.956 & 0.928 & 0.924 & 0.946 & 0.876 & \textbf{0.871} & 0.914 & \textbf{0.869} & \textbf{0.970} & 0.919 & 0.913 & 0.925 & 0.940 &  0.910\\
& Ours & \textbf{0.957} & \textbf{0.930} & \textbf{0.925} & \textbf{0.948} & \textbf{0.877} & \textbf{0.871} & \textbf{0.916} & \textbf{0.869} & \textbf{0.970} & \textbf{0.920} & \textbf{0.914} & \textbf{0.926} & \textbf{0.941} & \textbf{0.920} \\
\midrule
\multirow{4}{*}{LPIPS$\downarrow$}
& DVR & 0.095 & 0.129 & 0.125 & 0.098 & 0.173 & 0.150 & 0.172 & 0.170 & 0.094 & 0.119 & 0.139 & 0.110 & 0.116 & 0.130\\
& SRN & 0.111 & 0.150 & 0.147 & 0.115 & 0.152 & 0.197 & 0.210 & 0.178 & 0.111 & 0.129 & 0.135 & 0.165 & 0.134 & 0.139 \\
& pixelNeRF & 0.084 & 0.116 & 0.105 & 0.095 & 0.146 & 0.129 & 0.114 & 0.141 & 0.066 & 0.116 & 0.098 & 0.097 & 0.111 & 0.108\\
& Ours & \textbf{0.061} & \textbf{0.080} & \textbf{0.076} & \textbf{0.085} & \textbf{0.103} & \textbf{0.105} & \textbf{0.091} & \textbf{0.116} & \textbf{0.048} & \textbf{0.081} & \textbf{0.071} & \textbf{0.080} & \textbf{0.094} & \textbf{0.082}\\
\bottomrule
\end{tabular}
}
\end{small}
\end{adjustbox}
\end{center}
\vskip -0.12in
\end{table*}

\subsubsection{Unseen-Category View Synthesis}

In order to evaluate how our model generalizes to categories not seen during training, we follow the settings in pixelNeRF, and use only three object categories for training, namely airplane, car, and chair. We then evaluate on 10 unseen object categories. Table \ref{tb:NMR_datasets_unseen} compares the performance of our method with several baselines. We achieve state-of-the-art performance in SSIM and LPIPS, while performing slightly worse than pixelNeRF \cite{pixelNeRF} on PSNR. Fig. \ref{fig:multi_cat_unseen} indicates that our method is able to learn a good object prior, allowing it to generate feasible geometry for unseen categories such as benches and sofas. We also observe that the novel view images predicted by pixelNeRF are consistently more blurry, which explains its better performance on PSNR. On contrary, our model predicts sharp images that are more favorable by human perception, resulting in better metrics like LPIPS.

\begin{table*}[!t]
\caption{\textbf{Quantitative results on unseen-category view synthesis.} Our model obtains slightly worse PSNR, similar SSIM and better LPIPS metrics when compared to pixelNeRF.}
\vskip -0.2in
\label{tb:NMR_datasets_unseen}
\begin{center}
\begin{adjustbox}{width=0.85\linewidth}
\begin{small}
{
\begin{tabular}{cccccccccccccccc}
\toprule
Metrics & Methods & bench & cbnt.  & disp. & lamp & spkr. & rifle & sofa & table & phone & boat & mean\\
\midrule
\multirow{4}{*}{PSNR$\uparrow$}
& DVR & 18.37 & 17.19 & 14.33 & 18.48 & 16.09 & 20.28 & 18.62 & 16.20 & 16.84 & 22.43 & 17.72 \\
& SRN & 18.71 & 17.04 & 15.06 & 19.26 & 17.06 & 23.12 & 18.76 & 17.35 & 15.66 & 24.97 & 18.71 \\
& pixelNeRF & \textbf{23.79} & \textbf{22.85} & \textbf{18.09} & \textbf{22.76} & \textbf{21.22} & \textbf{23.68} & \textbf{24.62} & \textbf{21.65} & \textbf{21.05} & \textbf{26.55} & \textbf{22.71}\\
& Ours       & 23.10 & 22.27 & 17.01 & 22.15 & 20.76 & 23.22 & 24.20 & 20.54 & 19.59 & 25.77 & 21.90 \\
\midrule
\multirow{4}{*}{SSIM$\uparrow$}
& DVR & 0.754 & 0.686 & 0.601 & 0.749 & 0.657 & 0.858 & 0.755 & 0.644 & 0.731 & 0.857 & 0.716\\
& SRN & 0.702 & 0.626 & 0.577 & 0.685 & 0.633 & 0.875 & 0.702 & 0.617 & 0.635 & 0.875 & 0.684\\
& pixelNeRF & 0.863 & 0.814 & \textbf{0.687} & 0.818 & 0.778 & 0.899 & 0.866 & \textbf{0.798} & \textbf{0.801} & 0.896 & \textbf{0.825}\\
& Ours & \textbf{0.865} & \textbf{0.819} & 0.686 & \textbf{0.822} & \textbf{0.785} & \textbf{0.902} & \textbf{0.872} & 0.792 & 0.796 & \textbf{0.898} & \textbf{0.825} \\
\midrule
\multirow{4}{*}{LPIPS$\downarrow$}
& DVR & 0.219 & 0.257 & 0.306 & 0.259 & 0.266 & 0.158 & 0.196 & 0.280 & 0.245 & 0.152 & 0.240\\
& SRN & 0.282 & 0.314 & 0.333 & 0.321 & 0.289 & 0.175 & 0.248 & 0.315 & 0.324 & 0.163 & 0.280 \\
& pixelNeRF & 0.164 & 0.186 & 0.271 & 0.208 & 0.203 & 0.141 & 0.157 & 0.188 & 0.207 & 0.148 & 0.182\\
& Ours & \textbf{0.135} & \textbf{0.156} & \textbf{0.237} & \textbf{0.175} & \textbf{0.173} & \textbf{0.117} & \textbf{0.123} & \textbf{0.152} & \textbf{0.176} & \textbf{0.128} & \textbf{0.150}\\
\bottomrule
\end{tabular}
}
\end{small}
\end{adjustbox}
\end{center}
\end{table*}

\subsection{3D Reconstruction}
\label{sec:3D_recon}

We now turn to the task of evaluating the 3D geometry learned by our approach in a self-supervised manner by minimizing a novel view synthesis objective. In this setting, we evaluate 3D reconstruction by taking the mean intersection-over-union (mIoU) over the predicted $\alpha$ volume (the last channel of $\mathbf{V}_s$) and the corresponding ground truth occupancy volume. We compare our model to several unsupervised 3D reconstruction methods: PrGAN \cite{progan}, PlatonicGAN/3D \cite{platonicgan}, Multi.-View \cite{yan2016perspective}, and 3DGAN\cite{wu2016learning}. PlatonicGAN and PrGAN adopt a adversarial approach to learn 3D reconstruction given a single image with a canonical view. For this evaluation, we utilize the model trained with category-agnostic supervision and report results on the \textit{airplane} class as introduced in \cite{platonicgan}. The predicted alpha volume is binarized using a threshold $\tau=0.05$. The ground truth data is obtained from the ShapeNet voxelized volumes \cite{3dr2n2} and upsampled from $32^3$ to $64^3$ via nearest-neighbor interpolation. We then calculate the mIoU score and report in Table \ref{tb:3D_datasets}. Results of other models are directly taken from PlatonicGAN \cite{platonicgan}. 

\begin{figure*}[t]
\begin{center}
\centerline{\includegraphics[width=1.0\linewidth]{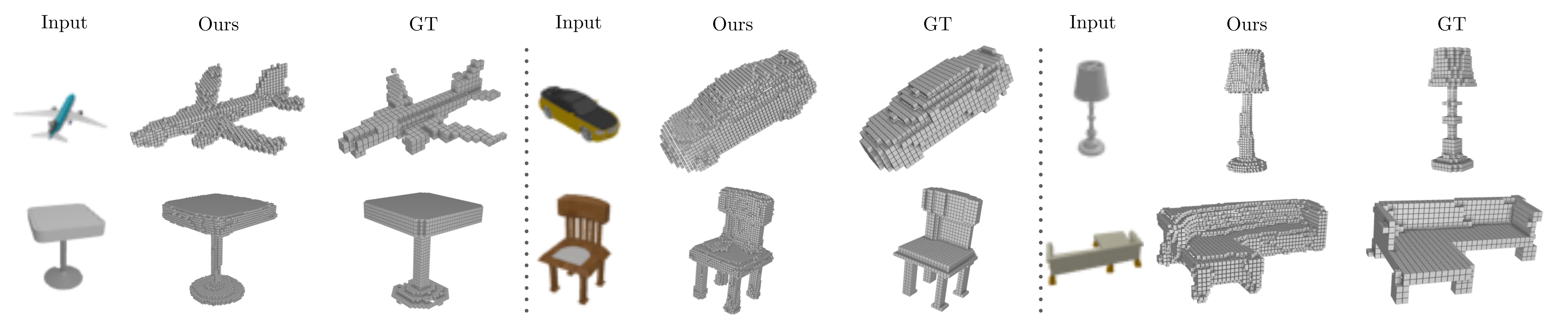}}
\caption{\textbf{Qualitative results for 3D geometry reconstruction.} We visualize the predicted $\alpha$ volume with its raw resolution $64^3$ and the ground truth volume with its raw resolution $32^3$. With a single-forward pass, our model can perform 3D geometry reconstruction given a single view of objects from 13 different categories. Our model trained with only 2D supervision consistently predicts meaningful and full geometry.}
\label{fig:3D_voxel_grid_visualization}
\end{center}
\vskip -0.1in
\end{figure*}

As shown in Table \ref{tb:3D_datasets}, our model predicts accurate 3D reconstruction, outperforming the best baseline by $10\%$ in mIoU. We attribute this boost in performance to the fact that our model can easlily tap large quantities of data in a category-agnostic manner. Whereas in GAN approaches like PlatonicGAN category-agnostic training has traditionally been a very an extremely difficult problem, preventing these approaches to tap large quantities of data for view synthesis. Fig. \ref{fig:3D_voxel_grid_visualization} shows qualitative 3D reconstruction results where we observe that our model produces accurate 3D models of objects. Furthermore, we extend the 3D reconstruction evaluation by including two \textbf{supervised} baselines V-LSMs\cite{vlsm} and 3D-R2N2 \cite{3dr2n2}. Our model obtains a mIoU of 63.25\% averaged across categories, while V-LSMs achieves 61.5\% and 3D-R2N2 achieves 55.1\%. Complete comparison details can be found in the appendix. 

\begin{table}[t]
\caption{\textbf{Quantitative results for 3D geometry reconstruction on Airplanes class.} Our model outperforms all baseline models in terms of single-view 3D reconstruction. }
\vskip -0.1in
\label{tb:3D_datasets}
\begin{center}
\begin{adjustbox}{width=1\linewidth}
\begin{small}
{
\begin{tabular}{ccccccccccc}
\toprule
& PrGAN & PlatonicGAN & Multi.-View & 3DGAN & PlatonicGAN 3D & Ours \\
\midrule
mIoU$\uparrow$ & 0.11 & 0.20 & 0.36 & 0.46 & 0.44 & \textbf{0.58} \\
\bottomrule
\end{tabular}
}
\end{small}
\end{adjustbox}
\end{center}
\end{table}

\begin{table}[!t]
\caption{\textbf{Ablation studies on different model components.} We show the effectiveness of various model components, trained with a $1/4$ size of the ShapeNet chairs dataset. }
\vskip -0.1in
\label{tb:ablation}
\begin{center}
\begin{adjustbox}{width=1\linewidth}
\begin{small}
{
\begin{tabular}{ccccccc}
\toprule
 & - inv projection & - 2D U-Net & - 3D U-Net & - half 3D resolution & Full\\
\midrule
PSNR & 20.62 & 21.10 & 21.45 & 21.82 & 21.94 \\
\bottomrule
\end{tabular}
}
\end{small}
\end{adjustbox}
\end{center}
\vskip -0.2in
\end{table}

\subsection{Ablation Studies}
\label{sec:design_ablation}

To better understand the benefits of each component of our model, we perform ablation studies by excluding one of each of the following components: inverse projection, 2D U-Net, 3D U-Net, or halved 3D voxel resolution. We use a $1/4$ training split of the ShapeNet chairs dataset and evaluate the performance on the full test split. Table \ref{tb:ablation} summarizes our findings. Starting from the right-most column, we sequentially remove and replace the components with their simplified variants and measure the model performance using the PSNR metric. It turns out that each component contributes \textit{[0.1, 0.5]} metric improvements. The inverse projection component is essential in terms of preserving the implicit geometric and texture information, in comparison to naively reshaping 2D feature volume into 3D \cite{enr, tbn}. 2D/3D U-Nets are useful to synthesize abstract geometry while preserving texture with skip connections, in comparison to single-path ResNet network structure. The halved 3D resolution is beneficial in reducing the tensor memory footprint and increasing the receptive field.

\section{Conclusion}
We have presented a simple yet effective approach to perform novel view synthesis of objects without explicit 3D supervision. Contrary to recent developments using radiance fields for view synthesis, our approach is neither continuous nor implicit. Despite the simplicity of our approach, we demonstrate that our model obtains comparable or even better performance than recent state-of-the-art approaches for few shot view synthesis using radiance fields \cite{pixelNeRF}, while rendering objects at over $400$x speed up. In addition, our model learns accurate 3D geometry in a self-supervised manner, relaxing the need of a large amount of 3D geometry data, and surpassing recent baselines for unsupervised learning of 3D geometry. 

As a future work (appendix), we plan to investigate the use of explicit sparse space representations such as octrees \cite{ogn, deepOCN, plenoctrees}, mixture of volumetric primitives \cite{mvp}, and scene graphs \cite{neuralscenegraph, giraffe} to increase our geometric capacity. Our current model cannot produce view-dependent lighting effects. This limitation can be addressed with a more informative material representation and a shading model that incorporates view direction, lighting, and surface information. We can also utilize techniques such as spherical harmonics \cite{sphericalharmonic} or a learned multilayer perceptron (MLP) to synthesize the color with view-dependent specular effects. By doing so during rendering time, we can leverage more advanced rendering techniques such as deferred rendering to better estimate the radiance field that captures both incoming light and material properties.

\newpage

{\small
\bibliographystyle{ieee_fullname}
\bibliography{egbib}
}

\clearpage

\appendix

\section{Limitations and Future Work}
Though our model shows very promising reconstruction results and great efficiency, a few limitations still exist to be tackled as future work. 

First, training a high fidelity as well as scene/category-agnostic representation remains an open problem for both explicit and implicit methods. The resolution of images that can be rendered by our model is capped by the resolution of the explicit voxel-like scene representation. Different from our approach, continuous representations like radiance fields are resolution-free by definition; however, they are either slow to obtain or slow to render. As a future work, we are planning to tackle this problem by increasing scene representation capacity for complex regions of space while minimizing the computational resources for empty regions. This can be achieved via a more flexible and sparse explicit space representations including octrees \cite{ogn, deepOCN, plenoctrees}, mixture of volumetric primitives \cite{mvp}, and scene graph \cite{neuralscenegraph, giraffe}.
In addition, it is also possible to enhance our model with super-resolution methods \cite{esrgan, esrgan+}.


Second, our current model cannot produce view-dependent lighting effects. We leave this as future work to be tackled with a physically based renderer that utilizes a more informative scene representation that incorporates view direction, light sources, material properties, and surface information \cite{nerd, nerfactor, nerv}. Specifically, we can utilize the Lambertian RGBA model as basis to form an albedo map, and accumulate additional view dependent lighting onto separate surface representations including material properties such roughness, metalness, and surface properties such as normals and displacement maps. Additionally, we can also leverage techniques such as spherical harmonics \cite{sphericalharmonic} or a learned MLP to synthesize the color with view-dependent specular effects. By doing so during rendering time, we can leverage more advanced rendering techniques such as deferred rendering to better estimate the radiance field that captures both incoming light and material properties.

\section{Social Impact}
Our work represents a step towards producing highly realistic generative models of the world. Such a goal can have both positive and negative social impacts. On the positive side, our model can enable interactive art creation, improved AR/VR experiences, etc. On the negative side, better generative models of 3D objects can potentially be misused to produce technology similar to deep fakes that have become a concern for misrepresenting person identity. Releasing code, models, and technical papers can help the community prepare for this kind of misuse by detecting fake content or ensuring content is certified.

\begin{table*}[!t]
\caption{\textbf{Quantitative results on unseen-category view synthesis.} Our model bypasses ENR \cite{enr} baseline by a noticeable margin. Our two-view model has a very large improvement from single-view counterpart, which indicates the good generalizability of the model.}
\label{tb:NMR_datasets_unseen_all}
\begin{center}
\begin{adjustbox}{width=1\textwidth}
\begin{small}
{
\begin{tabular}{cccccccccccccccc}
\toprule
Metrics & Methods & bench & cbnt.  & disp. & lamp & spkr. & rifle & sofa & table* & phone & boat & mean\\
\midrule
\multirow{3}{*}{PSNR$\uparrow$}
& ENR       & 22.59 & 21.41 & 16.99 & 22.29 & 20.13 & 23.22 & 23.16 & 20.00 & 20.15 & 25.81 & 21.50 \\
& Ours (Single-View)       & 23.10 & 22.27 & 17.01 & 22.15 & 20.76 & 23.22 & 24.20 & 20.54 & 19.59 & 25.77 & 21.90 \\
& Ours (Two-View)       & 25.11 & 24.54 & 20.85 & 25.19 & 22.72 & 27.74 & 26.24 & 23.34 & 23.80 & 29.17 & 24.80 \\
\midrule
\multirow{3}{*}{SSIM$\uparrow$}
& ENR & 0.845 & 0.799 & 0.674 & 0.819 & 0.768 & 0.900 & 0.848 & 0.759 & 0.806 & 0.896 & 0.807 \\
& Ours (Single-View) & 0.865 & 0.819 & 0.686 & 0.822 & 0.785 & 0.902 & 0.872 & 0.792 & 0.796 & 0.898 & 0.825 \\
& Ours (Two-View) & 0.909 & 0.875 & 0.818 & 0.905 & 0.838 & 0.963 & 0.913 & 0.871 & 0.897 & 0.950 & 0.893 \\
\midrule
\multirow{3}{*}{LPIPS$\downarrow$}
& ENR & 0.182 & 0.199 & 0.273 & 0.203 & 0.202 & 0.143 & 0.166 & 0.206 & 0.182 & 0.154 & 0.190\\
& Ours (Single-View)& 0.135 & 0.156 & 0.237 & 0.175 & 0.173 & 0.117 & 0.123 & 0.152 & 0.176 & 0.128 & 0.150\\
& Ours (Two-View) & 0.108 & 0.122 & 0.153 & 0.118 & 0.140 & 0.072 & 0.098 & 0.106 & 0.107 & 0.087 & 0.107\\
\bottomrule
\end{tabular}
}
\end{small}
\end{adjustbox}
\end{center}
\vskip -0.1in
\end{table*}

\section{Experiment Details}
\subsection{Implementation Details}
The 2D feature encoder utilizes the U-Net \cite{unet} with a ResNet-18 \cite{resnet} backbone, initialized with ImageNet pre-trained weights. The intermediate feature channels are $64, 128, 256, 512$ at each spatial resolution level during downsampling. At the upsampling stage, we perform deconvolution and fuse the resulting features with skip encoder features at the same level through concatenation followed by two consecutive convolution blocks. The final 2D feature map has the same spatial resolution as the input image, with a 32 channel feature map. 

At the inverse projection stage, the 2D feature map is back-projected into a 3D voxel space that has halved spatial height and width with respect to the input feature map. To enable back-projection, we first tile the 2D features into a 3D cube to prepare for sampling the inverse projection. Next, we leverage the camera intrinsic parameters to compute the homography matrix to map the cube to the viewing frustum. The grid mapping function is obtained by multiplying the homography matrix with individual points from the fixed voxel mesh grid set. Lastly, we apply the $\textit{grid\_sample}$ function to perform the inverse projection. 

The 3D feature decoder consists of a 3D U-Net with ResNet-3D blocks. Following common practice, when spatial resolution is downscaled by 2 times, we double the channel numbers. The output 3D feature map has 16 channels with the same spatial resolution as the input image. At the final layer, we apply one 3D convolution to transform the tensor map into 4 channels and sigmoid activations to produce the RGB$\alpha$ volume. 

At the amortized rendering stage, we combine the rotation transformation and the perspective deformation into one by left matrix multiplication, where rotation transformation transfers a volume from the source view to the target view and perspective deformation maps the viewing frustum to the output cube. After similar mesh grid multiplication and grid sampling steps as in inverse projection, we obtain the voxel volume used for alpha blending. The alpha blending process leverages $\textit{cumprod}$ method to calculate the accumulated transmittance between the near plane and the current point. 

We use the ReLU nonlinearity and GroupNorm \cite{groupnorm} normalization throughout the model for each nonlinear transformation. We train the network for 150 epochs with Adam optimizer and a fixed learning rate of $0.0016$, the loss weighting for SSIM \cite{ssim} loss is set to be $0.05$. We only perform data normalization with ImageNet statistics at the preprocessing stage. 

\subsection{Computation Resources}
For the category-specific view synthesis, each model is trained using 30 V100 GPUs on an internal cluster. It takes 5 days to train the single-chair dataset, 2.5 days to train the single-car dataset, and 1.5 days to train the multiple-chair dataset.

For the category-agnostic and unseen-category view synthesis, each model is trained using 30 V100 GPUs on an internal cluster, and it takes us 2.5 days to train each dataset.

We are able to perform the evaluation on a single GPU in minutes compared to PixelNeRF\cite{pixelNeRF}, which takes days for evaluation.

\begin{figure*}[t]
\begin{center}
\centerline{\includegraphics[width=0.85\linewidth]{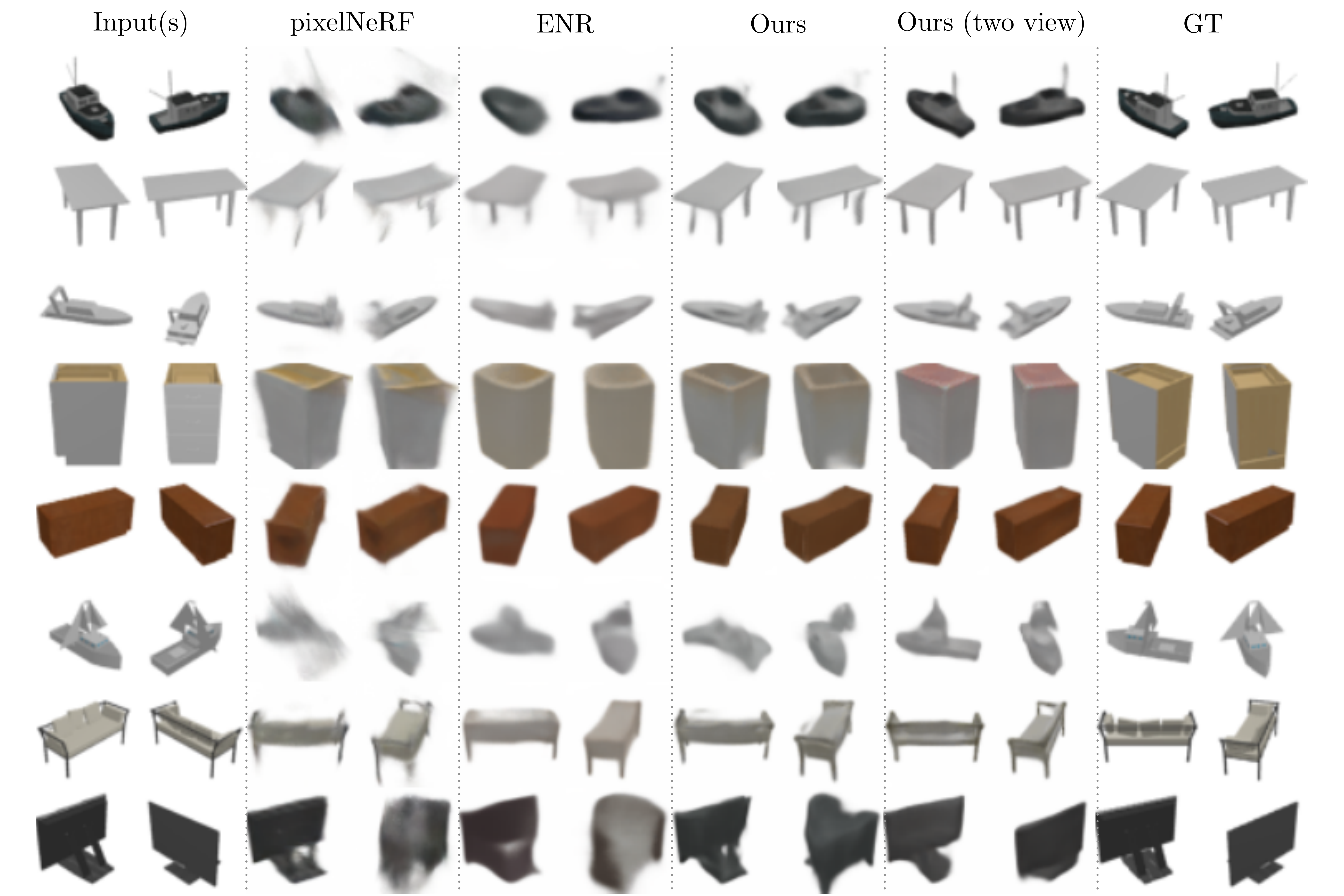}}
\caption{\textbf{Qualitative results on unseen-category datasets.} All methods except ours (two view) tasks a single input view (left image under Input(s) column) and perform novel view synthesis.}
\label{fig:multi_cat_unseen_all}
\end{center}
\vskip -0.3in
\end{figure*}

\begin{table*}[t]
\caption{\textbf{Quantitative results for 3D geometry reconstruction on 13 ShapeNet classes.} Our model has better mIoU metric compared to V-LSMs\cite{vlsm} and 3D-R2N2 \cite{3dr2n2} (1-view) that leverage groudtruth voxel occupancies as the supervision signal. }
\label{tb:3D_recon_depth}
\begin{center}
\begin{adjustbox}{width=1\textwidth}
\begin{small}
{
\begin{tabular}{cccccccccccccccc}
\toprule
Metrics & Methods & plane & bench & cbnt. & car & chair & disp. & lamp & spkr. & rifle & sofa & table* & phone & boat & mean\\
\midrule
\multirow{2}{*}{mIoU$\uparrow$}
& 3D-R2N2 w/pose \cite{3dr2n2} (1 view) & 56.7 & 43.2 & 61.8 & 77.6 & 50.9 & 44.0 & 40.0 & 56.7 & 56.5 & 58.9 & 51.6 & 65.6 & 53.1 & 55.1\\
& V-LSMs\cite{vlsm} (1 view) & \textbf{61.1} & 50.8 & 65.9 & 79.3 & \textbf{57.8} & 53.9 & 48.1 & 63.9 & \textbf{69.7} & 67.0 & \textbf{55.6} & 67.7 & 58.3 & 61.5\\
& Ours (1 view)       & 57.7 & \textbf{54.7} & \textbf{76.0} & \textbf{80.4} & 57.0 & \textbf{60.6} & \textbf{51.8} & \textbf{74.1} & 60.2 & \textbf{72.3} & 53.8 & \textbf{72.0} & \textbf{60.0} & \textbf{63.25} \\
\bottomrule
\end{tabular}
}
\end{small}
\end{adjustbox}
\end{center}
\vskip -0.1in
\end{table*}

\section{Additional Experiments \& Visualizations}
\subsection{Additional Experiments for Unseen-Category View Synthesis}
To further demonstrate the generalization capacity of our method, we extend the baselines for unseen-category view synthesis experiments to include ENR\cite{enr}. Our proposed approach beats ENR \cite{enr} in all metrics as shown in Table \ref{tb:NMR_datasets_unseen_all}. Figure \ref{fig:multi_cat_unseen_all} also shows that our model is able to synthesize objects with more clean geometry (less blurry artifacts ) as seen from the visualizations of boat (row 1), bench (row 7) objects. In addition, we also report results for a two-view model. During training, we randomly sample pairs of source images as input; during evaluation, we sample two images offset by 90 degrees azimuth as input. Figure \ref{fig:multi_cat_unseen_all} shows this model is not biased towards the training categories anymore, instead, the model is learning to perform view synthesis given two views of an unseen object.

\subsection{Additional Experiments for 3D Reconstruction}

We further demonstrate the 3D reconstruction performance of our model by comparing with two \textbf{supervised} 3D reconstruction baselines V-LSMs\cite{vlsm} and 3D-R2N2 \cite{3dr2n2}. These baselines use voxel occupancy as supervision, while our model only relies on 2D self-supervision. To conduct a fair comparison, we downsampled the predicted voxel from $64^3$ to $32^3$ and utilize the same thresholding strategy as reported in Section 4.2. Table \ref{tb:3D_recon_depth} indicates that our model obtains better mIoU on 9 out of 13 categories of objects and better mean mIoU across categories.

\subsection{Real World View Synthesis}
\begin{figure*}[t]

\begin{center}
\centerline{\includegraphics[width=1.0\textwidth]{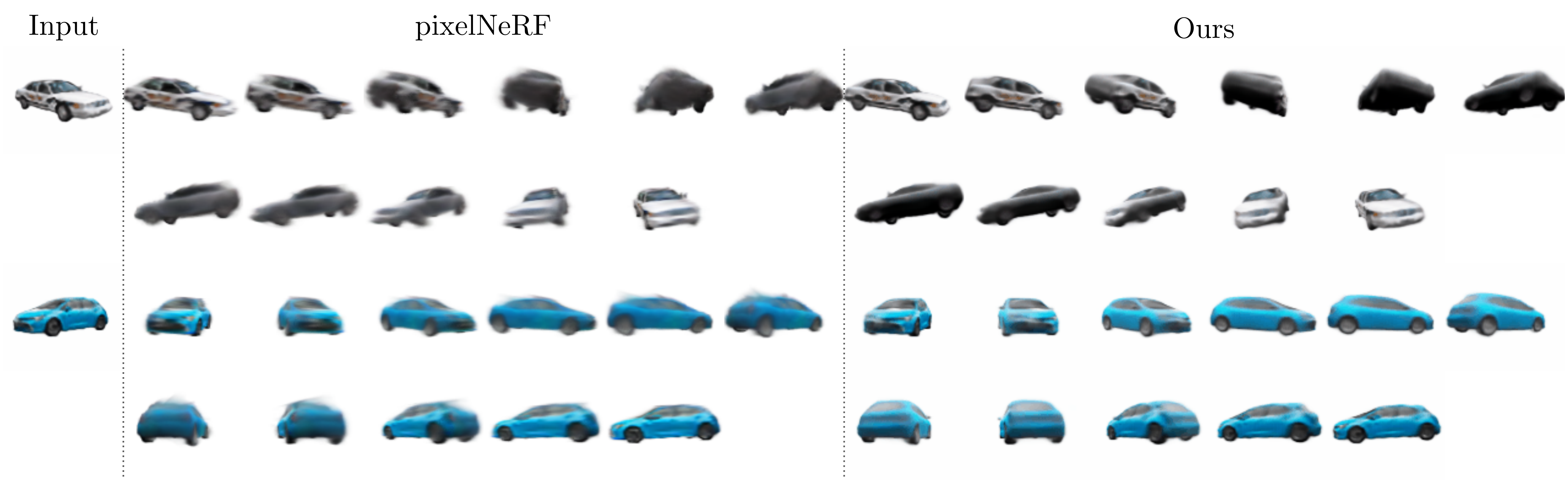}}
\caption{\textbf{Qualitative results for Real Car Image.} We use the model trained with ShapeNet virtual cars objects and perform novel view synthesis on real car images. Our model is able to synthesize plausible novel views, similar to pixelNeRF\cite{pixelNeRF} but with less blurry effects. }
\label{fig:real_car_vis}
\end{center}
\vskip -0.3in
\end{figure*}

We also test our model generalization capabilities when  transferring to a new target domain. We use the model trained with ShapeNet synthetic car objects from Section 4.1.1 of the main text and perform novel view synthesis on real car images from \cite{realcar}. Following the same protocol as pixelNeRF \cite{pixelNeRF}, we masked out the background and paint as white using PointRend\cite{pointrend}, and perform view synthesis in input view coordinate space. Figure \ref{fig:real_car_vis} suggests that our model can predict plausible novel views of the real cars. The geometry is similar to pixelNeRF\cite{pixelNeRF} results with less artifacts and blurry effects. 

\begin{table*}[t]
    \centering
    \caption{\textbf{Quantitative results on category-specific multiple chairs with increased number of views.} }
        \label{tb:multi_chair_increase}
        \vskip 0.2in
    \scalebox{0.95}{
    \centering
        \begin{tabular}{ccccccccccccccccccc}
    \toprule
    \multirow{2}{*}{Methods} & \multicolumn{3}{c}{1-view}  & \multicolumn{3}{c}{2-view}  & \multicolumn{3}{c}{3-view} \\
    \cmidrule(lr){2-4}
    \cmidrule(lr){5-7}
    \cmidrule(lr){8-10}
     & PSNR$\uparrow$ & SSIM$\uparrow$ & LPIPS $\downarrow$
     & PSNR$\uparrow$ & SSIM$\uparrow$ & LPIPS $\downarrow$
     & PSNR$\uparrow$ & SSIM$\uparrow$ & LPIPS $\downarrow$\\
    \midrule
    Ours & 17.02 & 0.747 & 0.241 & 24.13 & 0.907 & 0.098 & 26.18 & 0.935 & 0.076\\
    \bottomrule
    \end{tabular}}
\end{table*}

\subsection{Increasing Number of Source Views on Category-specific Multiple Chairs}

\begin{figure*}[ht]
\begin{center}
\centerline{\includegraphics[width=0.75\textwidth]{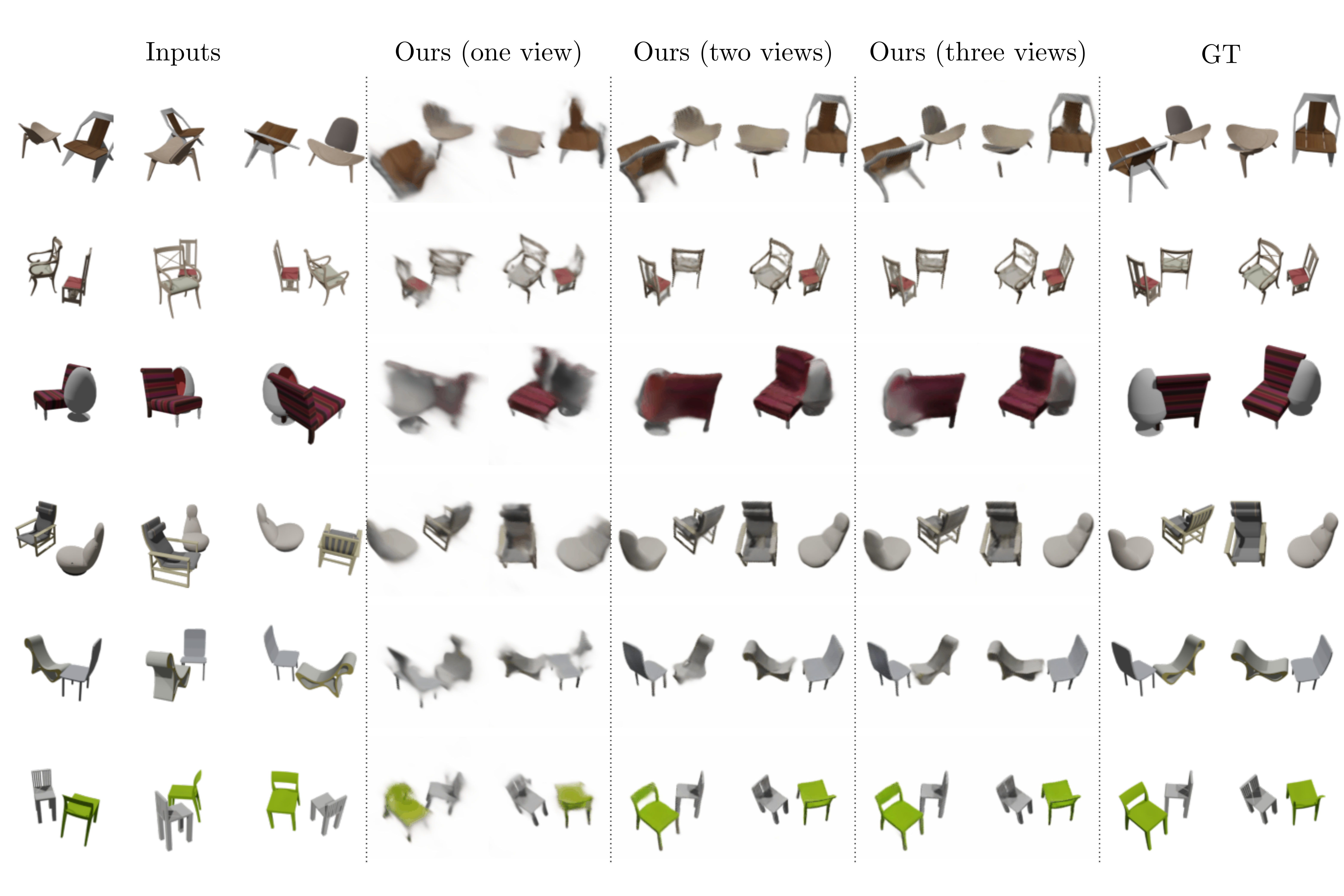}}
\vskip -0.1in
\caption{\textbf{Qualitative results on category-specific multiple chairs with an increased number of views.} We apply our model only trained with two input source viewers to a different number of input views settings during evaluation. Our model can predict better geometries and textures with the increasing number of views. }
\label{fig:multi_increase}
\end{center}
\vskip -0.25in
\end{figure*}

We analyze the capability of our model in terms of handling different numbers of source views. In this experiment, we use a model trained with two source views for evaluation. As shown in Table \ref{tb:multi_chair_increase} and Figure \ref{fig:multi_increase}, our model's performance decreases on the one view setting due to having to deal with a large number of degrees of freedom. As the number of source view increases, the model is able to satisfy more constraints; our model achieves the best performance in the three-view setting.

\subsection{Real World DTU dataset}

We evaluate the performance of our model in the real world DTU MVS dataset \cite{dtu}. As opposed to the ShapeNet\cite{shapenet} dataset, DTU contains a smaller number of scenes with very refined object textures, various lighting conditions and higher resolution images. This increased level of complexity requires our model to learn a good prior over shapes and textures to generalize across scenes given limited training samples. For this experiment, we follow the protocol in \cite{pixelNeRF} and use 88 training scenes and 15 test scenes. The images resolution is $300 \times 400$. During training, we randomly sample three images of the same object as input; during evaluation, we used a fixed set of input images. To better accommodate the scene in our voxel scene representation, we tune the physical size through the voxel distance between the object to camera used in inverse projection and perspective deformation steps. 

\begin{figure*}[ht]
\begin{center}
\centerline{\includegraphics[width=1.0\textwidth]{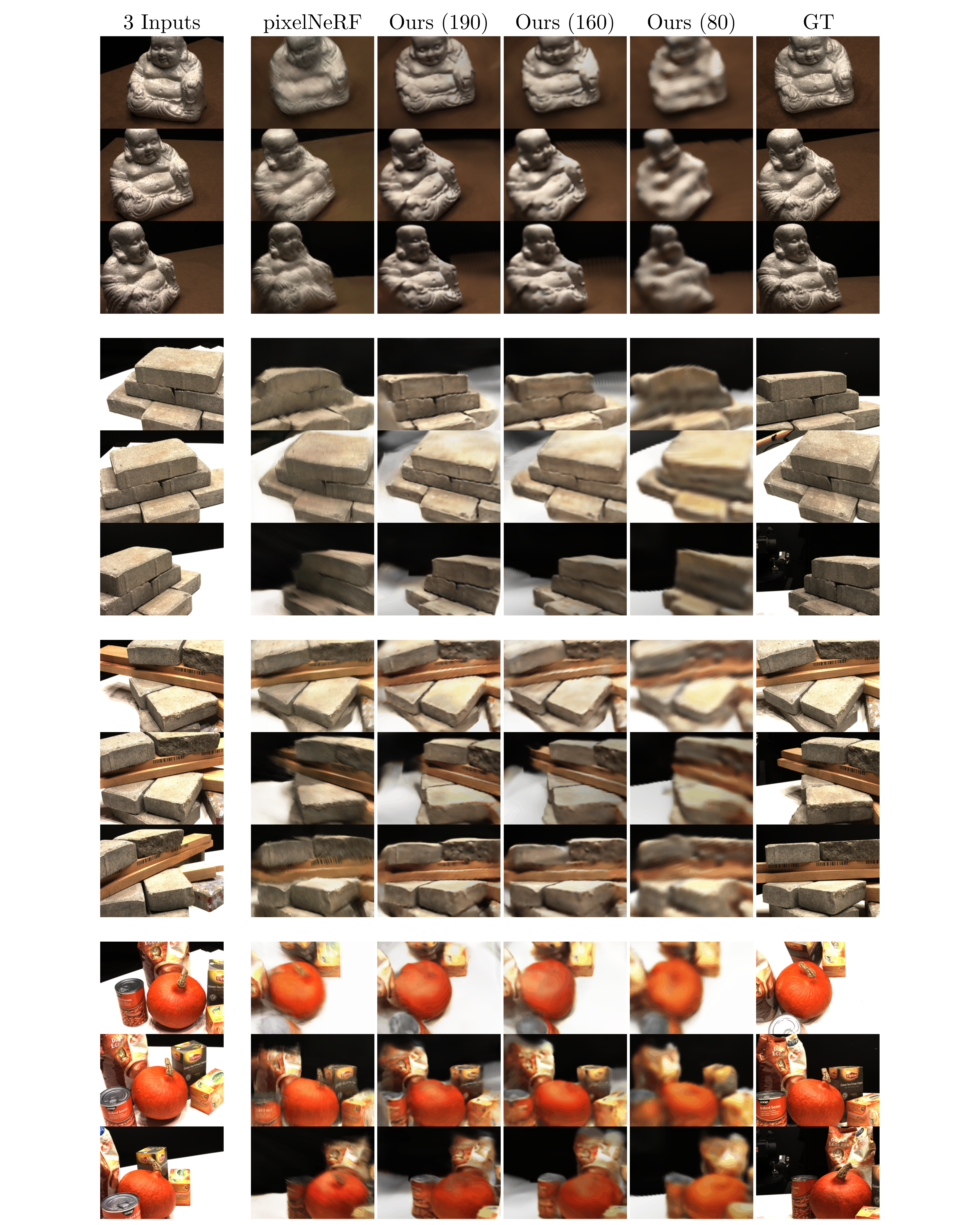}}
\vskip -0.1in
\caption{\textbf{Qualitative results on unseen test set scenes for DTU dataset.} Our model is able to synthesize reasonable novel views given 3 inputs of unseen scenes  while doing so at 1900x the speed of pixelNeRF.}
\label{fig:dtu}
\end{center}
\vskip -0.25in
\end{figure*}

As shown in Fig. \ref{fig:dtu}, our model is able to synthesize accurate novel views given 3 inputs of unseen scenes while doing so at 1900x the speed of pixelNeRF \cite{pixelNeRF}. With a larger voxel distance between the object to camera (190 v.s. 80), the physical size of the voxel scene representation is smaller, the object texture is more refined under the same voxel resolution. In that case, we are not able to estimate the background table correctly as the table is beyond the scope of the voxel representation. Literature like \cite{dvr} solve this problem by masking out the background and focus on reconstructing the foreground object. As a result, the reconstruction accuracy of our method is slightly impact when compared pixelNeRF as shown in Table \ref{tb:dtu}. Efficiency wise, as pixelNeRF's rendering time increases linearly with the resolution of the image, our model is able to increase per-view inference and rendering speed by over 600x, per-object inference and rendering speed by over 1900x. As discussed in the limitations section, an exciting future work to mitigate the small gap in performance could be to include a multi-resolution voxel representation \cite{ogn}.

\begin{table*}[t]
    \centering
    \caption{\textbf{Quantitative results on unseen test set scenes for DTU dataset.} 
    Our model sacrifices performance in the evaluation metrics mainly due to the limited physical size of the voxel scene representation. Meanwhile, it increases inference and rendering speed by 600x per-view wise and 1900x per object wise compared to pixelNeRF.
    }
        \label{tb:dtu}
        \vskip 0.2in
    \scalebox{0.95}{
    \centering
        \begin{tabular}{ccccccccccccccccccc}
    \toprule
    \multirow{2}{*}{Methods} & \multicolumn{3}{c}{3-view} & \multicolumn{2}{c}{Inference and Rendering Time (s)}  \\
    \cmidrule(lr){2-4}
    \cmidrule(lr){5-6}
     & PSNR$\uparrow$ & SSIM$\uparrow$ & LPIPS $\downarrow$ & Per-view & Per-object (46 views)
     \\
    \midrule
    NeRF (partial set) & 9.85 & 0.374 & 0.622 & - & -\\
    pixelNeRF (partial set) & 19.24 & 0.687 & 0.399 & - & -\\
    pixelNeRF & 18.99 & 0.680 & 0.420 & 35.9782 & 1655\\
    \midrule
    Ours (190) & 16.49 & 0.660 & 0.469 \\
    Ours (160) & 16.92 & 0.657 & 0.471 & 0.0557 & 0.8523\\
    Ours (80) & 17.58 & 0.645 & 0.494\\
    \bottomrule
    \end{tabular}}
\end{table*}

\subsection{Additional Visualizations}
\begin{figure*}[t]
\begin{center}
\centerline{\includegraphics[width=0.75\textwidth]{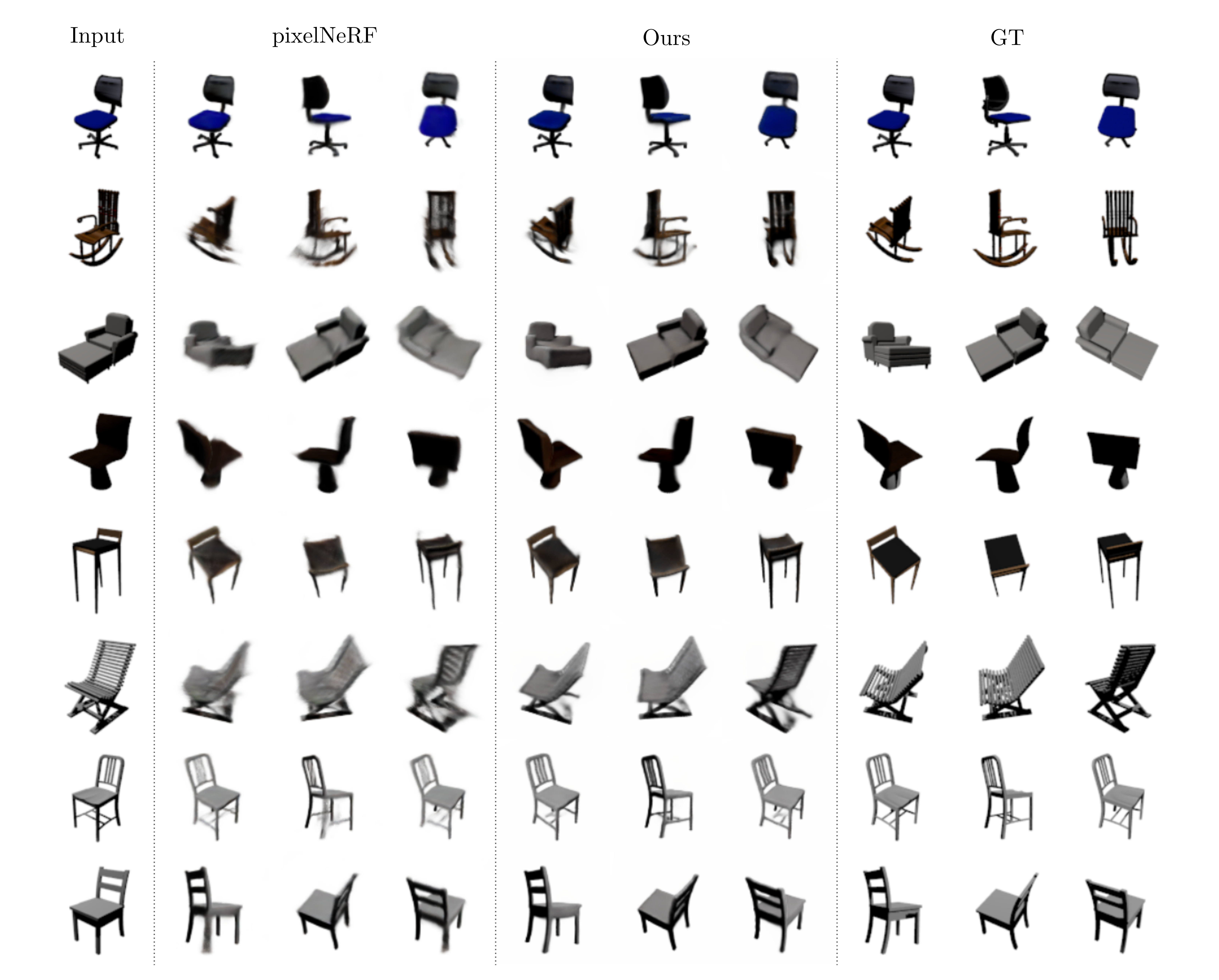}}
\vskip -0.1in
\caption{\textbf{Qualitative results on category-specific single chair (single-view)} }
\label{fig:chair_single_full}
\end{center}
\vskip -0.25in
\end{figure*}

\begin{figure*}[t]
\begin{center}
\centerline{\includegraphics[width=0.75\textwidth]{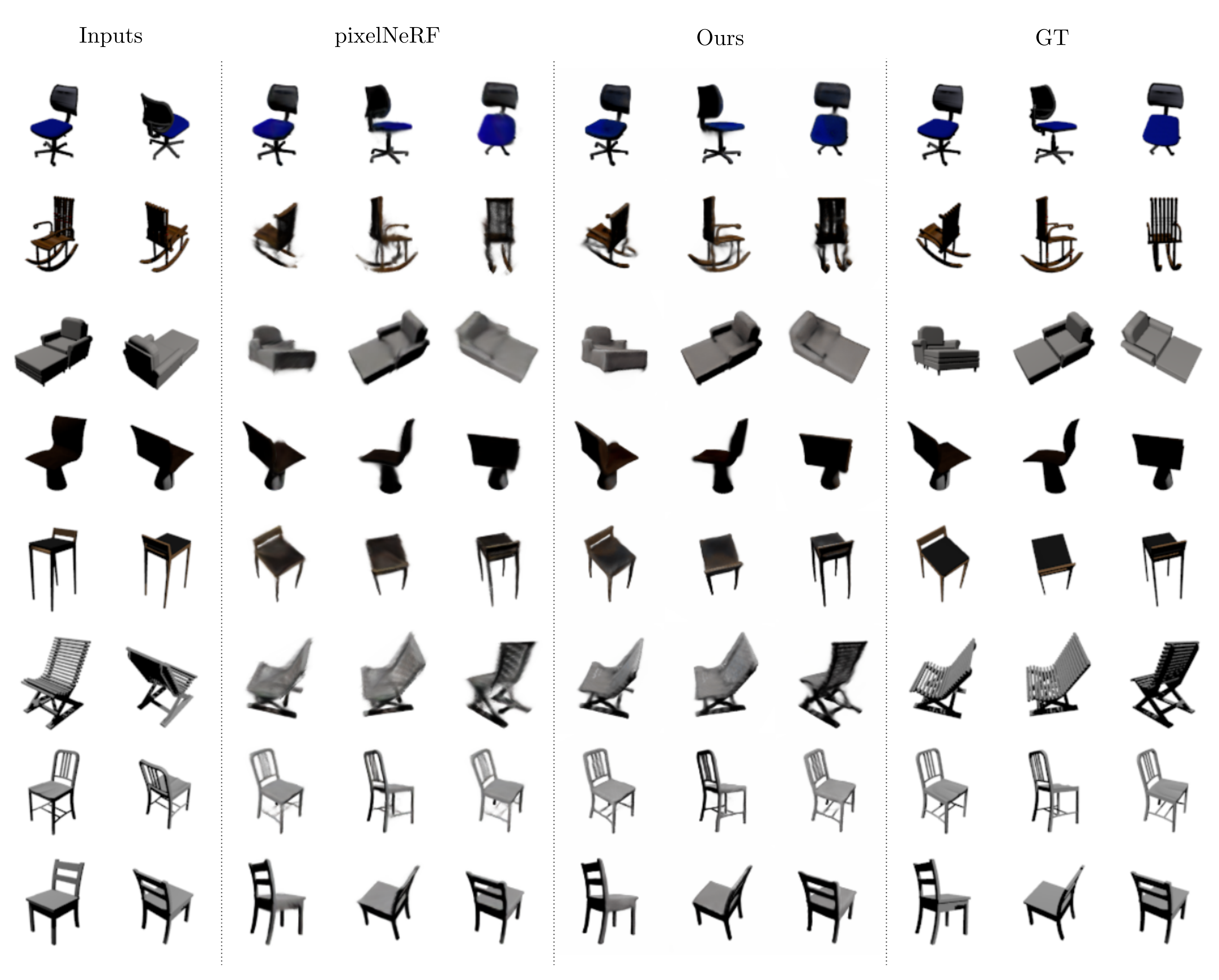}}
\vskip -0.1in
\caption{\textbf{Qualitative results on category-specific single chair (two-view)} }
\label{fig:chair_two_full}
\end{center}
\vskip -0.25in
\end{figure*}

\begin{figure*}[t]
\begin{center}
\centerline{\includegraphics[width=0.75\textwidth]{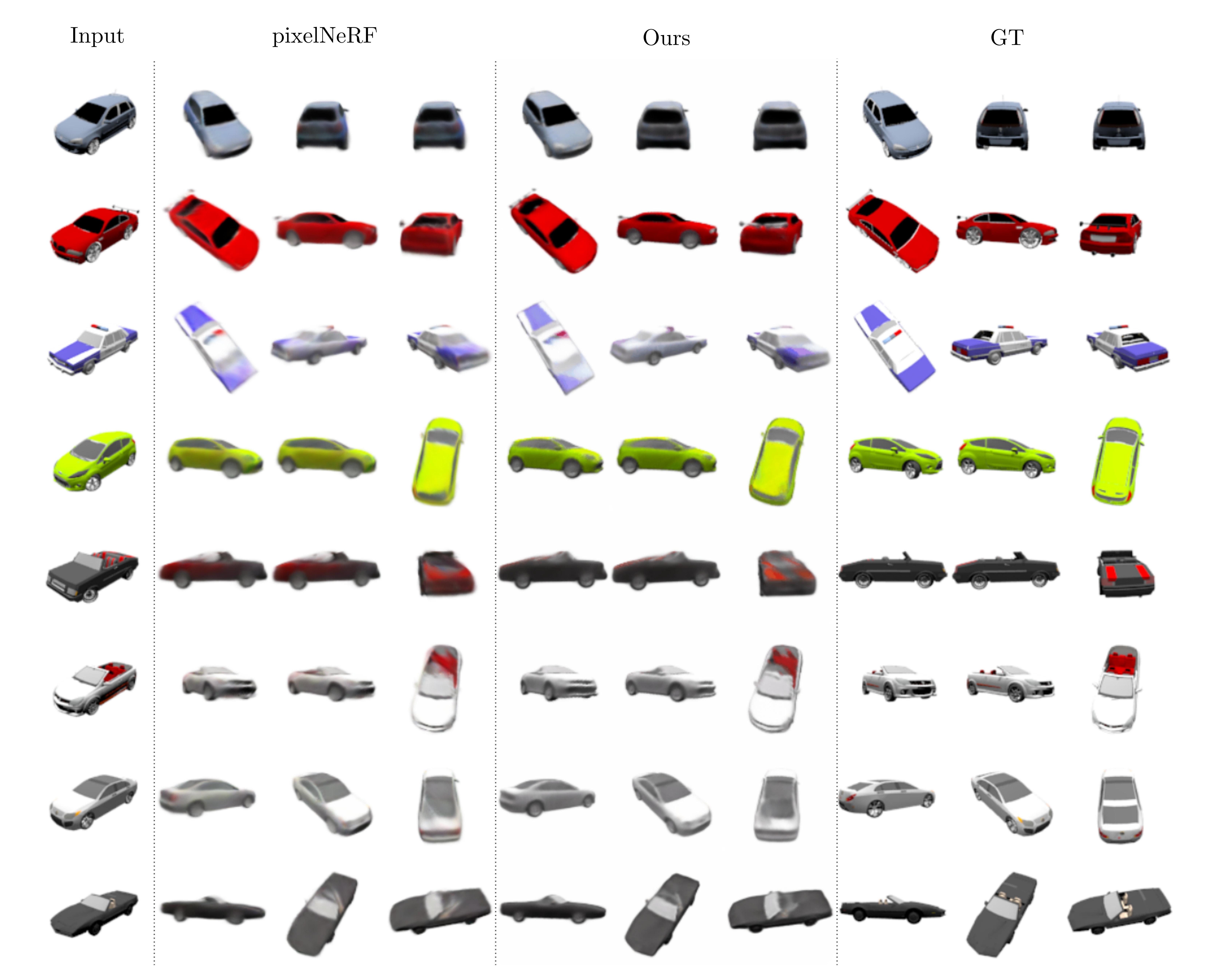}}
\vskip -0.1in
\caption{\textbf{Qualitative results on category-specific single car (single-view)} }
\label{fig:car_single_full}
\end{center}
\vskip -0.25in
\end{figure*}

\begin{figure*}[t]
\begin{center}
\centerline{\includegraphics[width=0.75\textwidth]{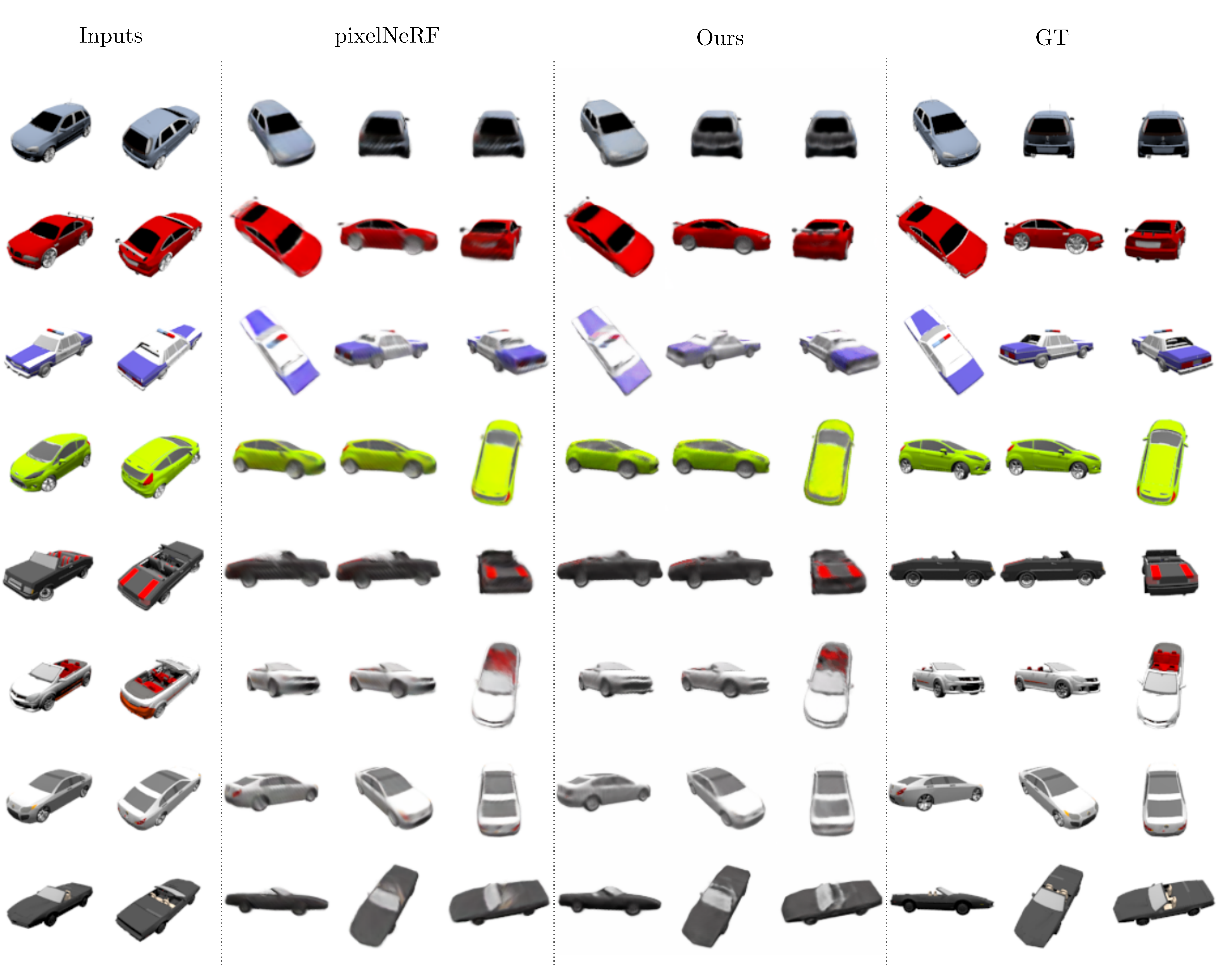}}
\vskip -0.1in
\caption{\textbf{Qualitative results on category-specific single car (two-view)} }
\label{fig:car_two_full}
\end{center}
\vskip -0.2in
\end{figure*}

\begin{figure*}[t]
\begin{center}
\centerline{\includegraphics[width=0.75\textwidth]{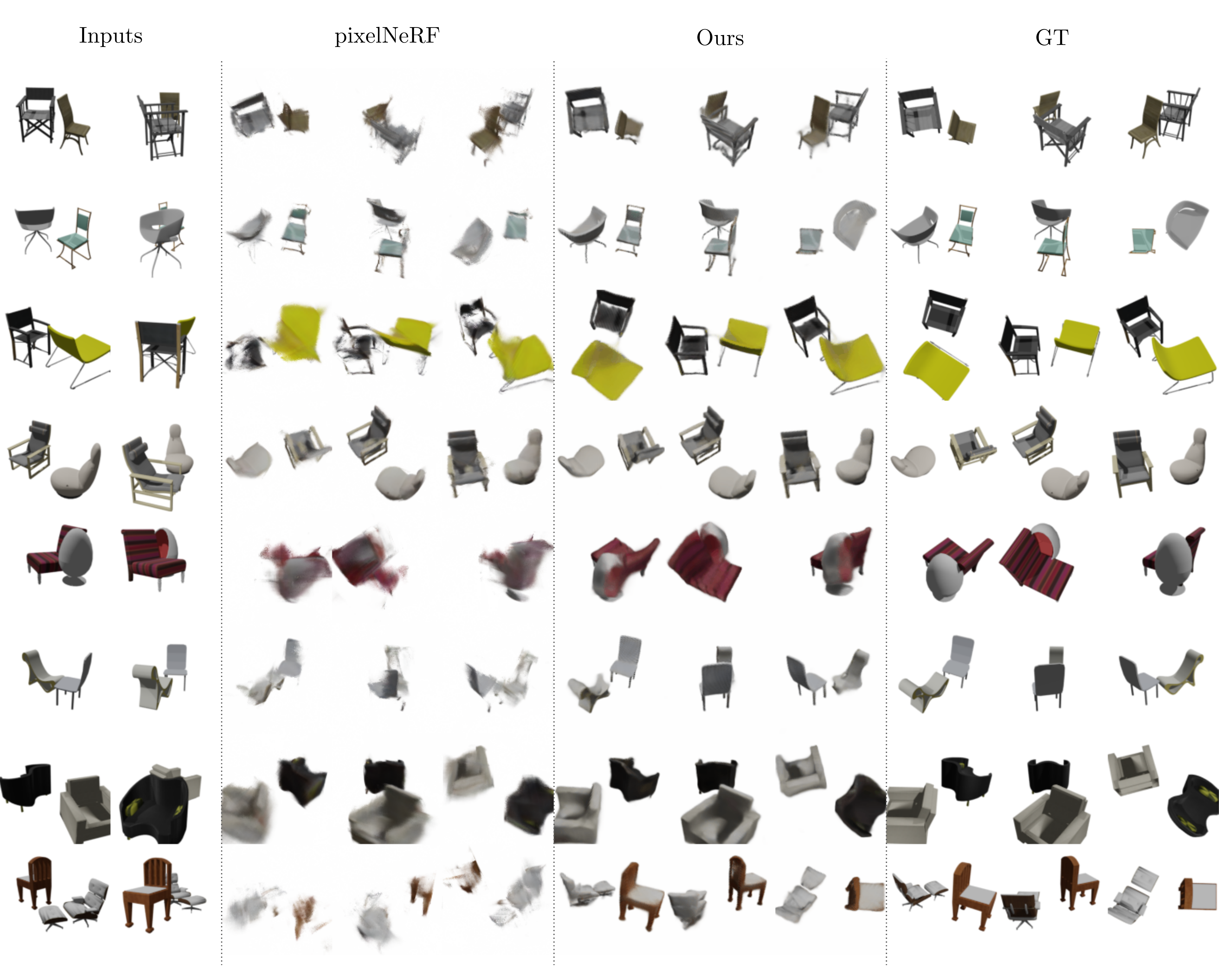}}
\vskip -0.1in
\caption{\textbf{Qualitative results on category-specific multiple chairs (two-view)} }
\label{fig:multi_chair_two_full}
\end{center}
\vskip -0.25in
\end{figure*}

\begin{figure*}[t]
\begin{center}
\centerline{\includegraphics[width=0.75\textwidth]{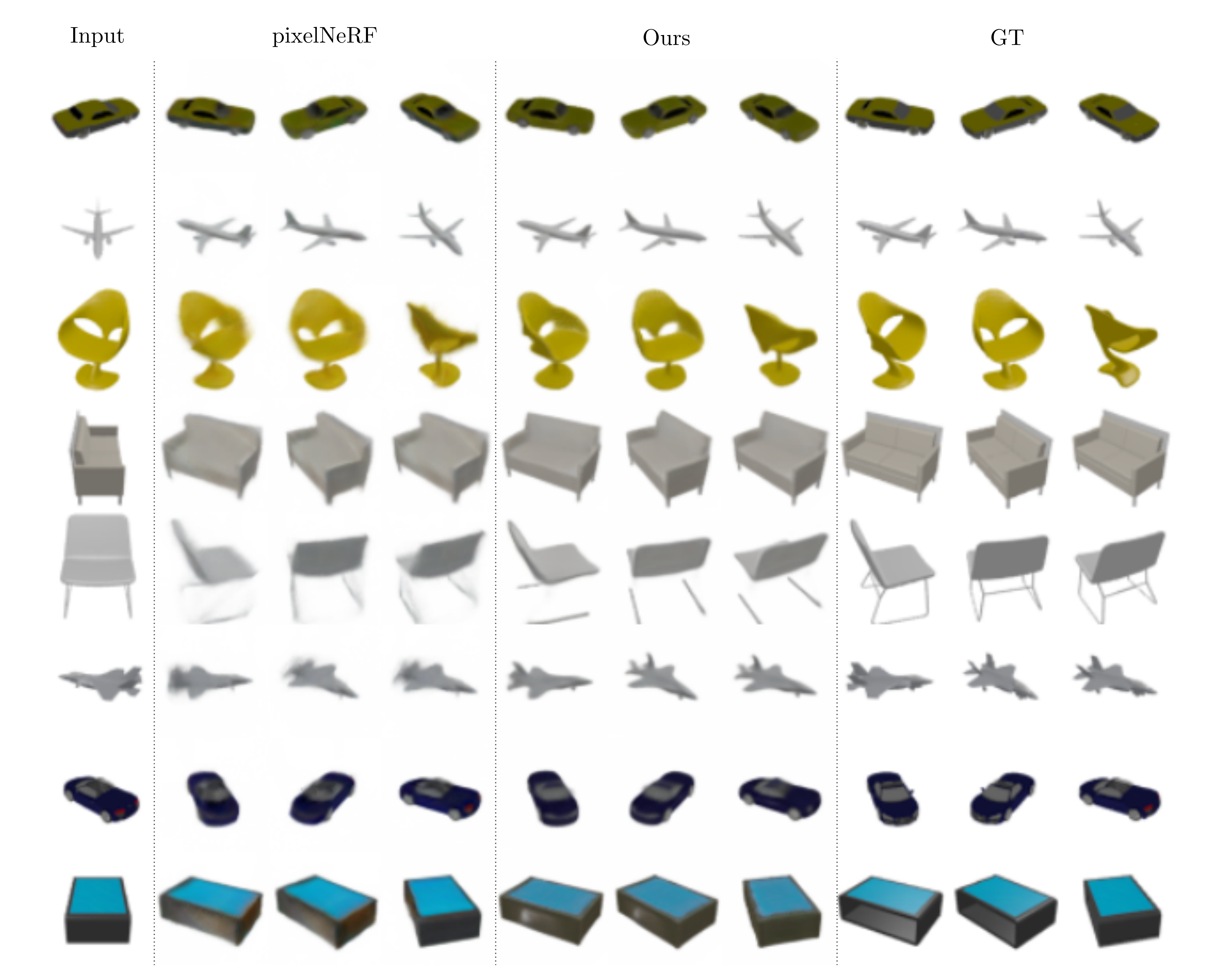}}
\vskip -0.1in
\caption{\textbf{Qualitative results on multiple-category dataset (one-view)} }
\label{fig:multi_cat_full}
\end{center}
\vskip -0.25in
\end{figure*}

\begin{figure*}[t]
\begin{center}
\centerline{\includegraphics[width=0.75\textwidth]{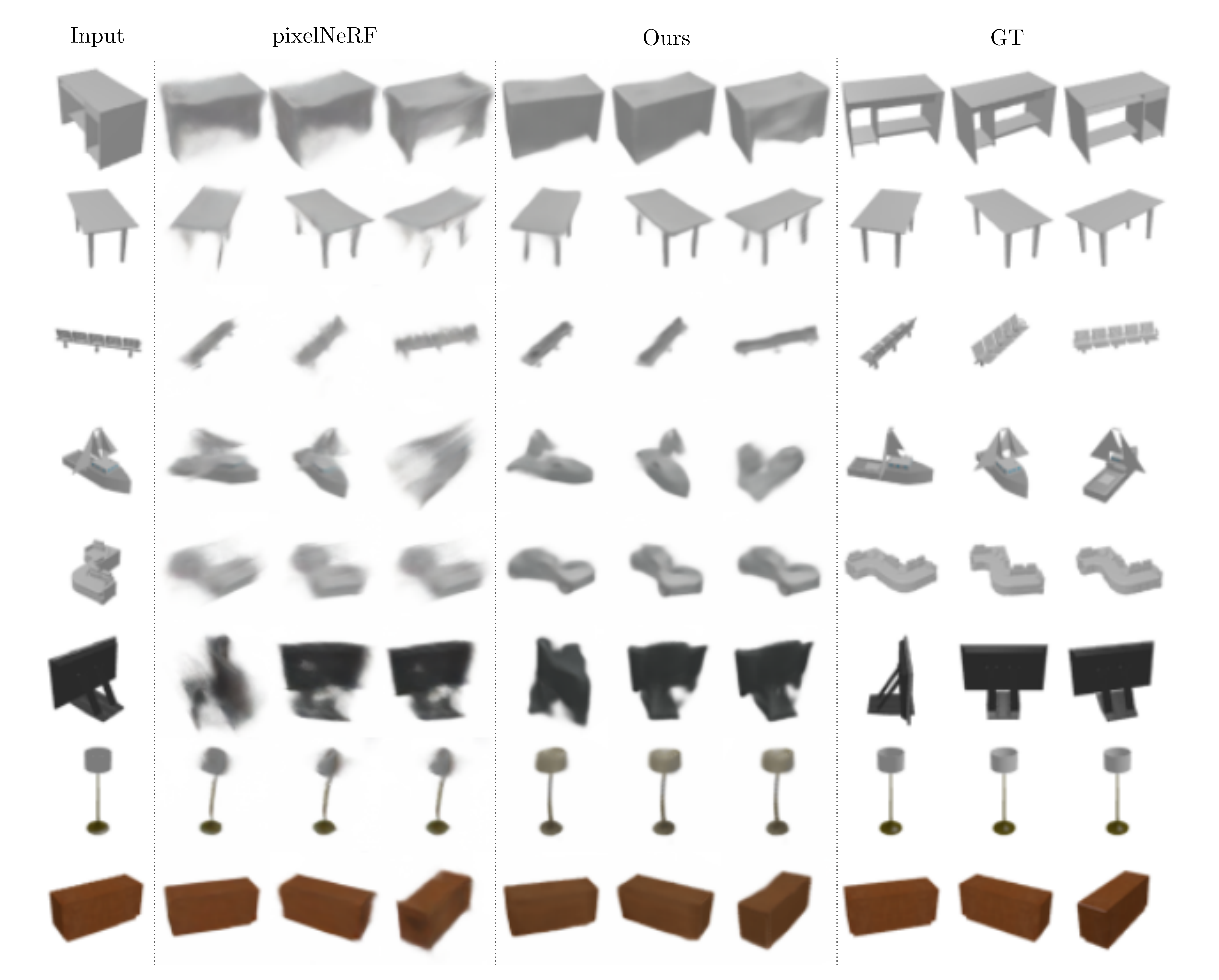}}
\vskip -0.1in
\caption{\textbf{Qualitative results on unseen-category dataset (one-view)} }
\label{fig:unseen_cat_full}
\end{center}
\vskip -0.25in
\end{figure*}

We include additional visualizations for every experiment:

\begin{itemize}
    \item Figure \ref{fig:chair_single_full}: category-specific single chair dataset with one source view;
    \item Figure \ref{fig:chair_two_full}: category-specific single chair dataset with two source views;
    \item Figure \ref{fig:car_single_full}: category-specific single car dataset with one source view;
    \item Figure \ref{fig:car_two_full}: category-specific single car dataset with two source views;
    \item Figure \ref{fig:multi_chair_two_full}: category-specific multiple chairs dataset;
    \item Figure \ref{fig:multi_cat_full}: category-agnostic dataset;
    \item Figure \ref{fig:unseen_cat_full}: unseen-category dataset.
\end{itemize} 

We intentionally use the same random indices for different view settings of the single chair/car dataset. From the visualizations in Figure \ref{fig:chair_single_full} v.s. Figure \ref{fig:chair_two_full} and Figure \ref{fig:car_single_full} v.s. Figure \ref{fig:car_two_full}, it can be easily observed that our model can synthesize better geometry and texture at novel target pose with more information given. 

\newpage

\end{document}